\pdfoutput=1 

\documentclass[final]{cvpr}

\usepackage{flushend}
\usepackage{times}
\usepackage{epsfig}
\usepackage{graphicx}
\usepackage{amsmath}
\usepackage{amssymb}
\usepackage{bm}
\usepackage{color} 
\usepackage{caption}
\usepackage{enumitem}
\usepackage{multirow,tabularx}
\usepackage{url}

\usepackage[pagebackref=true,breaklinks=true,colorlinks,bookmarks=false]{hyperref}

\usepackage[labelfont=bf]{caption} 

\definecolor{darkorange}{rgb}{1.0, 0.55, 0.0}

\def\cvprPaperID{5420} 
\def\confYear{CVPR 2021}

\title{HumanGAN: A Generative Model of Human Images} 
\author{Kripasindhu Sarkar \qquad Lingjie Liu \qquad Vladislav Golyanik \qquad Christian Theobalt\vspace{7pt}\\
Max Planck Institute for Informatics, SIC 

}
\begin{document}
\twocolumn[{%
\renewcommand\twocolumn[1][]{#1}%
\maketitle
\begin{center}
    \centering
    \includegraphics[width=1\textwidth]{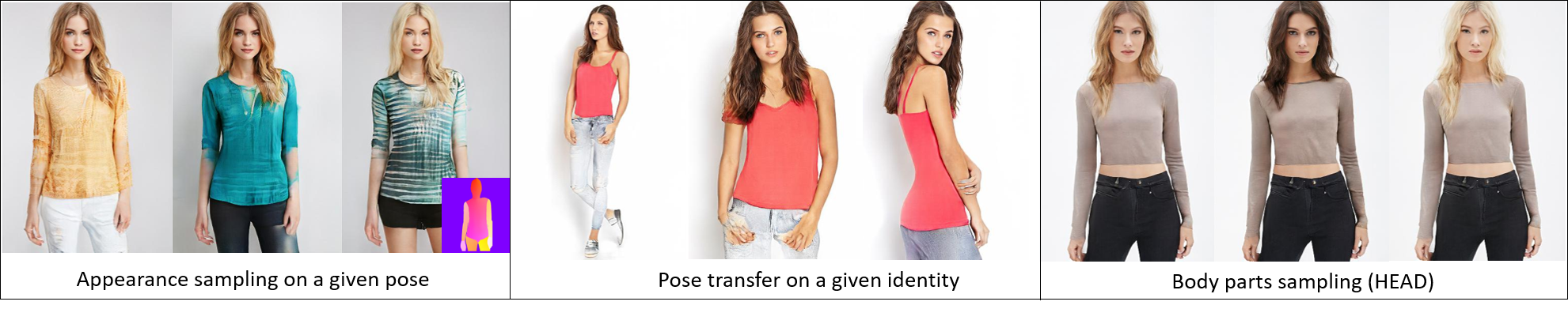} 
    \captionof{figure}{
    %
    %
    \textbf{Exemplary samples synthesized by \textit{HumanGAN}}.
    Our pose-guided generative model can sample random appearances conditioned on a given pose \textit{(left)},
    create persistent appearance and identity across different poses \textit{(middle)}, and sample body parts \textit{(right)}. 
    %
    %
    All the nine images shown in this figure are  synthesized by our generative model. 
    } 
    \label{fig:teaser} 
\end{center}%
}]

\begin{abstract}
Generative adversarial networks achieve great performance in photorealistic image synthesis in various domains, including human images. 
However, they usually employ latent vectors that encode the sampled outputs globally.
This does not allow convenient control 
of semantically-relevant individual parts of the image, and cannot draw  samples that only differ in partial aspects, such as clothing style. 
We address these limitations and present a generative model for images of dressed humans offering control over pose, local body part appearance and garment style. 
This is the first method to solve various aspects of human image generation, such as global appearance sampling, pose transfer, parts and garment transfer, and part sampling jointly in a unified framework. 
As our model encodes part-based latent appearance vectors in a normalized pose-independent space and warps them to different poses,  it preserves body and clothing appearance under varying posture. 
Experiments show that our flexible  and  general  generative method outperforms task-specific baselines for pose-conditioned image generation, pose transfer and part sampling in terms of realism and output resolution.
\end{abstract}

\section{Introduction}
\label{sec:intro}


Algorithms to generate images of clothed humans find many applications in such fields as virtual and augmented reality, data generation and augmentation for neural network training. 
%
%
For content creation, it is often desired to have full control over semantic properties of the generated images (\textit{e.g.,} pose, body appearance and garment style). 
One way of achieving this is to use computer graphics with precise control over the image rendering. 
However, creating just a single photo-realistic image in this way is challenging and tedious and requires expert knowledge in 3D modeling,  animation and rendering algorithms.

Recently, generative adversarial networks (GANs) have made significant progress in generating photo-realistic images, which can also be applied to synthesize imagery of humans 
\cite{Brock2018Biggan,karras2018progressive,karras2019style,Karras2019stylegan2}.  
These methods learn a mapping from an easy-to-sample latent space that can be sampled to the image domain and is able to synthesize photo-realistic imagery  without the need to resort to complex compute graphics style modeling and light transport simulation. 
A limitation of these generative models from a content creation perspective is that, in contrast to computer graphics synthesis, they do not easily permit control over  semantic attributes of the output imagery. 
Recently, Tewari \textit{et al.}~\cite{tewari2020stylerig,tewari2020pie} study StyleGAN \cite{karras2019style} to achieve a rig-like control over 3D interpretable face parameters, such as face pose, expression and illumination. 
However, these methods only work well for faces with limited 3D poses \cite{tewari2020stylerig}, and extending such a method to humans---that relies on differentiable rendering of a 3D morphable model---is not straightforward. 

In a setting related to ours, there has been  significant progress in conditional image generation with explicit inputs of specific control variables \cite{pix2pix2017, wang2018pix2pixHD}. 
Conditional models of this type that use conditioning inputs from a parametric human body model have shown impressive results in applications such as pose and garment transfer \cite{Sarkar2020,Grigorev2019CoordinateBasedTI,Siarohin2019AppearanceAP,Neverova2018}. 
Unfortunately, their underlying translation network does not constitute a full generative model that can be sampled from, as they are designed to produce a single output deterministically.

In this work, we present \textit{HumanGAN}, \textit{i.e.,} a novel generative model for full-body images of clothed humans, which enables control of  body pose, as well as independent control and sampling of  appearance and clothing style on a body part level \footnote{project webpage: \url{gvv.mpi-inf.mpg.de/projects/HumanGAN/}}. 
Our method combines the advantages of both worlds,  \textit{i.e.,} the latent-space-based GANs and controllable  translation networks, in the framework of conditional  variational autoencoder \cite{Kingma2014VAE}. 
We encode the true posterior probability of the latent variable from a space of pose-independent appearance and reconstruct a photo-realistic image of the human with a high-fidelity generator using the encoded latent vector. 
To disentangle a pose from local part appearance, we propose a novel strategy where we condition the latent vectors on body parts and warp them to a different pose before performing the reconstruction. 
This permits pose control under persistent appearance and clothing style, and appearance sampling on a localized body part level, without affecting the pose, see Fig.~\ref{fig:teaser}.
To summarize, our \textbf{contributions} are as follows: 
\vspace{-0.2cm}
\begin{itemize}[leftmargin=*]
\itemsep-0.2em
\item HumanGAN, \textit{i.e.,} a new state-of-the-art generalized model for human image generation that can perform global appearance sampling, pose transfer, parts and garment transfer, as well as part sampling. 
For the first time, a \textit{single method} can support \textit{all these tasks} (Sec.~\ref{sec:method}). 
%
\item A novel strategy of part-based encoding and part-specific spatial warping in a variational framework that disentangles pose and appearance over the body parts. 
\end{itemize}
\vspace{-0.2cm}
In our experiments (Sec.~\ref{sec:results}), we significantly outperform the state of the art (and other tested methods) for human appearance sampling in realism, diversity and output resolution ($512 \times 512$).
Furthermore, our general model shows commendable results for \textit{pose transfer} that are on par with the state-of-the-art methods developed for this task. 


\section{Related Work}

\noindent\textbf{Deep Gernerative Models} have made remarkable achievements on image generation. 
As the original GAN model~\cite{Goodfellow14} was only able to synthesize low-resolution images, 
the follow-ups improved it with multiple 
discriminators~\cite{GMAN,Mordido2018DropoutGANLF,Doan2019OnlineAC}, self-attention  mechanism~\cite{sagan,Brock2018Biggan} and progressive training  strategy~\cite{karras2018progressive}. 
These methods use a single latent vector $\mathbf{z}$ to resemble the latent factor distribution of training data, which leads to unavoidable entanglements  and limited control over image synthesis. 
StyleGANs~\cite{karras2019style, Karras2019stylegan2} 
approach this problem by mapping $\mathbf{z}$ to an intermediate latent space $\mathbf{w}$, which is then fed into the generator to change different levels of attributes. 
Although it enables 
more control on image synthesis, it does not disentangle different feature factors. 
Recent works~\cite{tewari2020stylerig,tewari2020pie} extend StyleGAN to synthesize images of faces with a rig-like control over 3D interpretable face parameters such as face pose, expression and scene illumination. 
Compared to faces, synthesizing the full human appearance with control of 3D body pose and human appearance is a much more difficult problem due to more severe 3D pose and appearance changes. 
We propose the first method to this problem allowing photo-realistic image synthesis of a full human body with controls of the 3D pose as well as the appearance of each body part. %

\noindent\textbf{Conditional GAN} 
(cGAN) 
uses conditional information for the generator and discriminator. cGAN is useful for applications such as class conditional image generation~\cite{mirza2014conditional,miyato2018cgans,pmlr-v70-odena17a} and image-to-image translation~\cite{pix2pix2017,wang2018pix2pixHD}.
Many works~\cite{mirza2014conditional,pix2pix2017,wang2018pix2pixHD,park2019SPADE,wang2018vid2vid} require paired data for fully-supervised training. 
Pix2Pix~\cite{pix2pix2017} and Pix2PixHD~\cite{wang2018pix2pixHD} learn the mapping from input images to output images. 
Some works~\cite{CycleGAN2017,YiZTG2017,LiuBK2017,choi2017stargan,Recycle-GAN} tackle a harder problem of learning the mapping between two domains based on unpaired data. 
cGAN is a deterministic model which produces a single output. 
Our approach also applies a conditional GAN to map from a warped noise image to output images. 
However, unlike cGAN methods, we are able to randomly sample noise from a normal distribution for each body part for synthesizing different output images.

\begin{figure*}[t] 
    \includegraphics[width=\linewidth]{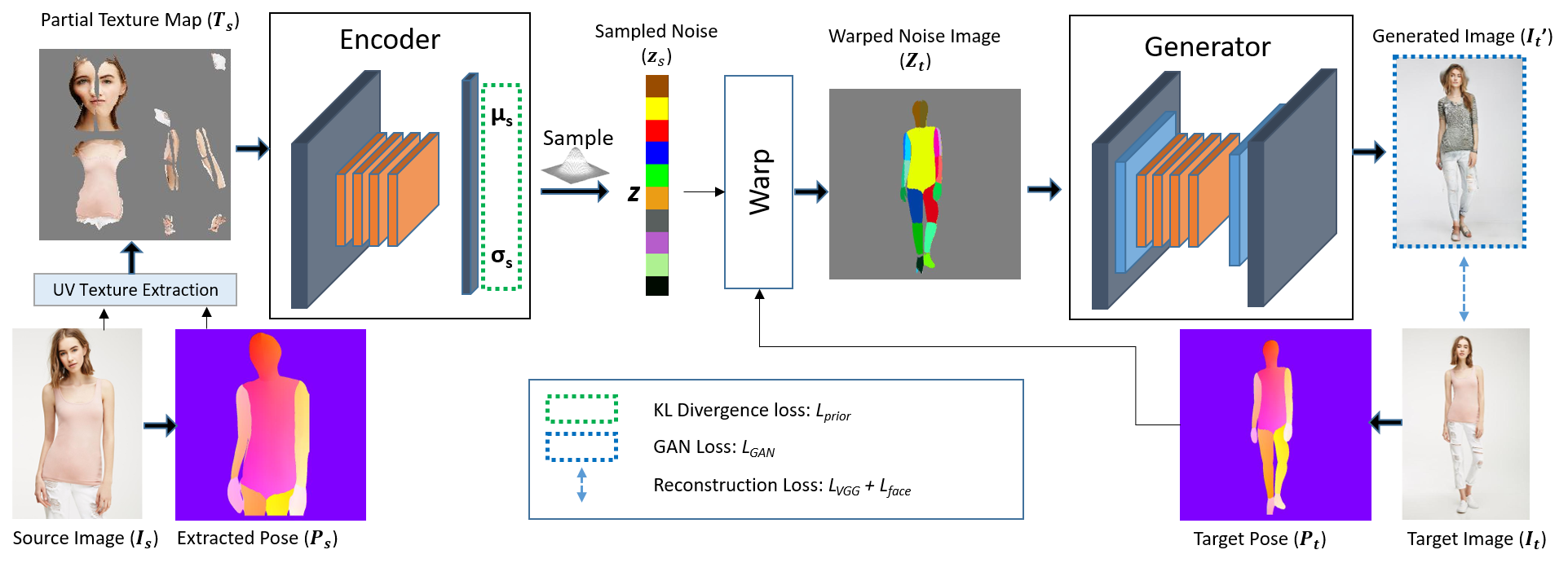} 
    \caption{\textbf{Overview of our method.} Given a source image $\bm I_s$, we extract a UV texture map of the human appearance $\bm T_s$. 
    The encoder then encodes $\bm T_s$ to part-specific latent vectors $\bm z_s$. 
    The target pose $\bm P_t$ is used to warp and broadcast the latent vectors to the corresponding parts in the target image to create a noise image $\bm Z_t$.
    Finally, the generator converts $\bm Z_t$ to a realistic image. 
    At testing, we generate random samples by controlling the latent vectors $\bf z \in \mathcal{N}(0,1)$ and the target pose $\bm P_t$.
    } 
    \label{fig:pipeline} 
    \vspace{-0.5cm}
\end{figure*}

\noindent\textbf{Pose Transfer} 
%
refers to the problem of transferring person appearance from one pose to another \cite{MaSJSTV2017}. 
Most approaches formulate it as an image-to-image mapping problem, \textit{i.e.,} given a reference image of target person, mapping the body pose in the format of renderings of a  skeleton~\cite{chan2019dance,SiaroSLS2017,Pumarola_2018_CVPR,KratzHPV2017,zhu2019progressive}, dense  mesh~\cite{Liu2019,wang2018vid2vid,liu2020NeuralHumanRendering,Sarkar2020,Neverova2018,Grigorev2019CoordinateBasedTI}, dense labeled pose maps \cite{Yoon2021} or joint position  heatmaps~\cite{MaSJSTV2017,Lischinski2018,Ma18} to real images. 
Though our method is not specifically designed for pose transfer, it can also be applied to this problem with high-quality results, as our generated samples retain identity across different poses. We demonstrate this in Sec.~\ref{sec:results}.

\noindent\textbf{Variational Autoencoders} 
 (VAE) are likelihood-based models which can effectively model stochasticity \cite{Kingma2014VAE}. 
Larsen \textit{et al.}~\cite{larsen2016autoencoding} first introduced a combined VAE and GAN to achieve higher quality than vanilla VAE. 
VUNet~\cite{esser2018variational} combined VAE with UNet for pose-guided image generation. 
Lassner \textit{et al.}~\cite{Lassner2017} present a two-stage VAE framework 
using a parametric model of human meshes \cite{SMPL:2015} as pose and shape conditioners. 
Our method builds on conditional VAE-GAN. 
Unlike the existing methods, it generates images of higher resolution and quality, and offers more control over part sampling and garment transfer. 
%


\section{Method}
\label{sec:method}

Our goal is to learn a generative model of human images, which is conditioned on body pose and a low dimensional latent vector encapsulating the appearance of different body parts. 
We use DensePose~\cite{Guler2018DensePose} to represent the human pose. Our task can be then formulated as a deterministic mapping $G:(\bm P, \bm z) \rightarrow \bm I$. Here, $\bm P \in \mathbb{R}^{H \times W \times 3}$ is a three-channel DensePose image representing the conditioning pose, $\bm z \in \mathbb{R}^{M \times N}$ is the latent vector comprised of $M$ human body parts and $N$ part-specific latent vector dimensions, and $\bm I \in \mathbb{R}^{H \times W \times 3}$ is the generated image.

For learning such function, we build our method on variational autoencoders (VAE) \cite{Kingma2014VAE}. 
As in any latent vector model, the observed variable of images $\bm I$ is assumed to be dependent on the unobserved hidden latent variable $\bm z$ that encodes semantically meaningful information about $\bm I$.  The goal is then to find their joint distribution $p(\bm{I, z})$ by maximizing the evidence lower bound (ELBO) -- by jointly optimizing the encoder $q(\bm{z|I})$ and the decoder $p(\bm{I|z})$.
In this work, our \textit{key assumption} is that the latent variable $\bm z$  depends \textit{only} on the appearance of the subject in the image $\bm I$. To enforce that, we first extract a pose-independent human appearance $\bm T$ from $\bm I$. Our encoder $q(\bm{z|T})$ is then conditioned on $\bm T$ (which is actually a function of $\bm I$), to encode to part-specific latent vector $\bm z$. 
Furthermore, the encoded appearance $\bm z$ is warped by a target pose $\bm P_t$ different from the pose in $\bm I$, which is subsequently used by a generator. 
We next describe our method in detail. 

\subsection{Our Architecture} 
In the training stage, we take pairs of images ($\bm{I_s}, \bm{I_t}$) of the same person (but in different poses) as input. Our method performs in four steps. 
In the first step,  we extract SMPL \textit{UV texture map} $\bm T_{s}$ from the input image $\bm I_s$  using the DensePose correspondences. In the second step, we use an encoding function $E$ to map the human appearance $\bm T_s$ of the source image to the parameters of the distribution of the latent vector. In the third step, we sample $\bm z$  from the estimated distribution of the source appearance. Given a target pose $\bm P_t$, we warp the encoded latent vector $\bm z$ to a noise image $\bm Z_t$. In the fourth step, we decode the warped $\bm Z_t$ to a realistic image $\bm I_t'$ by a high-fidelity generator network. Our method is summarized in Fig.~\ref{fig:pipeline}. 
 
\noindent\textbf{Extracting Appearance.} 
\label{sec:texmap}
We use a UV texture map of the SMPL surface model \cite{SMPL:2015} to represent the subject's appearance in the input image. 
The pixels of the input image $\bm{I_s}$ are transformed into the UV space through a mapping predicted by DensePose RCNN~\cite{Guler2018DensePose}. The pretrained network trained on COCO-DensePose dataset predicts  $24$ body segments and their part-specific \textit{U,V} coordinates of SMPL model. For easier mapping, the 24 part-specific UV maps are combined to form a single normalized UV texture map $\bm T_s$ in the format provided in SURREAL dataset \cite{varol17_surreal}. This normalized (partial) texture map provides us with a pose-independent appearance encoding of the subject that is located spatially according to the body parts. The 24 part segments in the texture map also provide us the placeholder for part-based noise sampling, \textit{i.e.,} in our case, the number of body parts $M = 24$. 

\noindent\textbf{Encoding Appearance.} 
\label{sec:encoding}
As with VAE, we assume the distribution $q(\bm{z|T_s})$ of the latent code $\bm z$, given the appearance $\bm T_s$ to be Gaussian, $q(\bm{z|T_s}) \equiv \mathcal{N}(\bm{\mu_s, \sigma_s})$. We use a convolutional neural network $E(\cdot)$ that takes the partial texture $\bm T_s$ as input and predicts the parameters ($\bm{\mu_s, \sigma_s}$) of the Gaussian distribution, $\bm{\mu_s, \sigma_s} \in \mathbb{R}^{M \times N}$. 
The encoder comprises a convolutional layer, five residual blocks, an average pooling layer, and finally, a fully connected layer that produces the final output.

\noindent\textbf{Warping Latent Space.} 
\label{sec:warping} 
In the next step, we sample a latent code from the predicted distribution of the encoded appearance, $\bm z_s \sim E(\bm T_s) \equiv \mathcal{N}(\bm{\mu_s, \sigma_s})$. 
Given the noise vector $\bm z_s$ and a target pose $\bm P_t$, 
we intend to reconstruct a realistic image $\bm I_t'$
with the appearance encoded in  $\bm z_s \in \mathbb{R}^{M \times N}$ and pose from $\bm P_t$. We also want the latent code for a specific body part to have direct influence on the same body part in the generated image. 
We enforce this by warping and broadcasting the part-based latent code to the corresponding part location in the target image and create a noise image $\bm Z_t \in \mathbb{R}^{H \times W \times N}$. Here, for each body part $k$, $\forall_{i,j \in k} \bm Z_t[i,j] \gets \bm z[k]$ (see the warping module in Fig.~\ref{fig:pipeline}). 
This operation can be easily implemented by differentiable sampling $W(\cdot)$ given the DensePose image $\bm P_t$, \textit{i.e.} $\bm Z_t = W(\bm z_s, \bm P_t)$. The design enables us to perform part-based sampling during the test time. 
Other straightforward ways of using $\bm z_s$, such as sampling noise in the UV texture space or tiling of a single noise vector in the entire spatial dimension, did not give us the required result. 
See Sec.~\ref{sec:res:appearance} and \ref{sec:res:parts} for a detailed analysis.

\noindent\textbf{Decoding to a Photo-Realistic Image.} 
\label{sec:decoding}
The warped noise image in the target pose $\bm Z_t$ with the noise vectors correctly aligned with the body parts in the target pose, is used as an input to a generator network $G(\cdot)$. 
The generator and the warping module act as the conditional decoder of our pipeline. 
We use the high-fidelity generator from Pix2PixHD \cite{wang2018pix2pixHD} that comprises three down-sampling blocks, six residual blocks and three up-sampling blocks. 
%


\subsection{Training Details}
\label{sec:training}

Our entire training pipeline can be summarized by the following equations: 
\begin{equation}\label{eq:sampling}
  \begin{aligned}
\bm z_s \sim E(\bm T_s), \quad \bm Z_t = W(\bm z_s, \bm P_t), \quad \bm{ I_{t}'} = G(\bm Z_t),\\
 \end{aligned}
\end{equation}

With the re-parameterization trick \cite{Kingma2014VAE} for sampling $\bm z_s$, the entire pipeline becomes differentiable, allowing direct back-propagation to the parameters of $E(\cdot)$ and $G(\cdot)$. The pipeline is trained with an objective $\mathcal{L}_{total}$ derived from conditional VAE-GAN: 
\begin{equation}\label{eq:loss}
 \mathcal{L}_{total} =  \mathcal{L}_{rec} +  \mathcal{L}_{GAN} +  \mathcal{L}_{prior}. 
\end{equation}
We describe in the following all three loss terms of \eqref{eq:loss}. 

\noindent\textbf{Reconstruction Loss.} 
The reconstruction loss $\mathcal{L}_{rec}= \lambda_{VGG} L_{VGG} + \lambda_{face}L_{face}$ quantifies the dissimilarity between the generated image $\bm {\tilde I_t}$ and the ground-truth image $\bm I_t$. 
It comprises 1) \textit{Perceptual Loss $L_{VGG}$} which is the difference between the activations on different layers of the pre-trained VGG network \cite{simonyan2014very} applied on the generated image and the ground-truth image; 
and
2) \textit{Face Identity Loss $L_{face}$}  which is the difference between features of the pre-trained SphereFaceNet \cite{liu2017sphereface} on the cropped 
face of the generated image and the ground-truth image.

\noindent\textbf{GAN Loss.} The GAN loss $\mathcal{L}_{GAN}= \lambda_{D} L_D + \lambda_{FM}L_{FM}$ pushes the generator to generate realistic images. We directly use the two-scale discriminator architecture $D$ from Pix2PixHD \cite{wang2018pix2pixHD} for implementing the GAN loss. The network $D$ is conditioned on both generated image and warped noise image at different scales. The total GAN loss  comprises of multiscale adversarial loss $L_D$ and discriminator feature matching loss $L_{FM}$. 
See \cite{wang2018pix2pixHD} for more details.

\noindent\textbf{Prior Loss $\mathcal{L}_{prior}$.} 
To enable sampling at inference time, the encoding $E(\bm T_s)$ is encouraged to be close to a standard Gaussian distribution, \textit{i.e.,} the prior distribution on the $\bm z$ 
vector is assumed to be $\mathcal{N}(0, I)$. Therefore, we employ the prior loss 
$\mathcal{L}_{prior} = \lambda_{KL} \mathcal{D}_{KL}(E(\bm T_s)||\mathcal{N}(0, I)),$
where $\mathcal{D}_{KL}(p||q)$ is the Kullback-Leibler divergence between the probability distributions $p(x)$ and $q(x)$. 

With reparameterization trick on sampling $\bm z_s$, we train the system end-to-end and optimize the parameters of the networks $E, G$ and $D$. The final objective $\mathcal{L}_{total}$ in Eq.~\eqref{eq:loss} is minimized with respect to the generator $G$ and the encoder $E$, while maximized with respect to the the discriminator $D$.
We use Adam optimiser \cite{adam} for our optimization with an initial learning rate of 2$\times 10^{-4}$, $\beta_1$ as 0.5 and no weight decay. The loss weights are set empirically to $\lambda_{VGG} = 10, \lambda_{face} = 5, \lambda_{D} = 1, \lambda_{FM} = 10, \lambda_{KL} = 0.01$. For speed, we pre-compute DensePose on the images and directly read them as input.

\begin{figure*}[t] 
    \includegraphics[width=\linewidth]{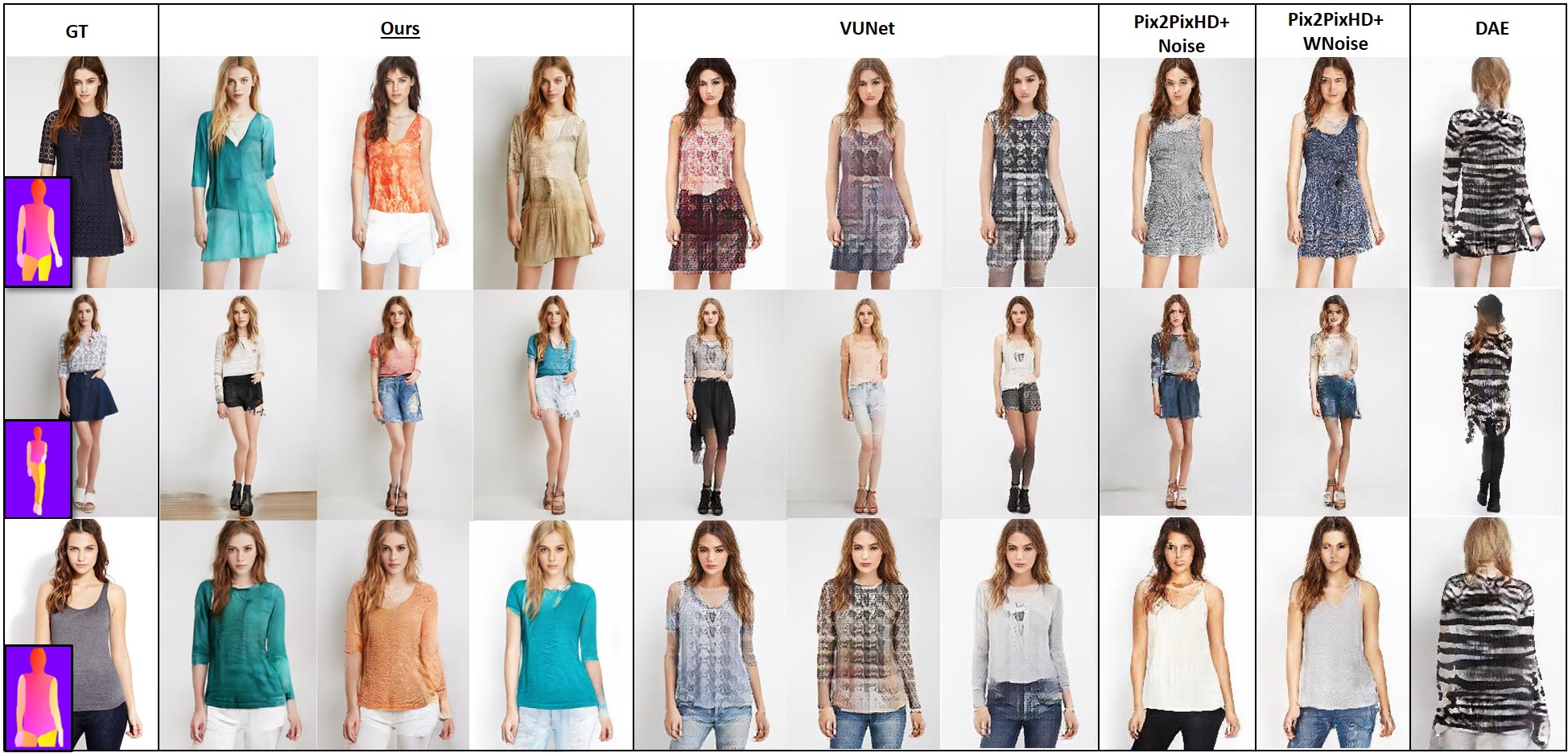} 
    \caption{Results of our HumanGAN approach for pose-guided image generation, and its comparison with VUNet \cite{esser2018variational}, Pix2PixHD+Noise, Pix2PixHD+WNoise and Deterministic Auto Encoder (DAE). The conditioning pose is shown in the left corner. Our approach produces samples of higher quality than the baseline methods.} 
    \label{fig:appearance} 
\end{figure*}

\subsection{Inference: Sampling Poses and Body Parts} 

During testing, we \textit{sample} the \textit{appearance} vector $\bm z$ from the prior distribution. We warp $\bm z$ with the conditioning pose $\bm P$ and feed the resulting noise image to the trained generator $G$ to get a generated image $\bm I_{z,P}$: 
\begin{equation}\label{eq:sampling}
  \begin{aligned}
\bm z \sim \mathcal{N}(0, I),  \quad \bm Z_P = W(\bm z, \bm P), \quad \bm{I_{z,P}} = G(\bm Z_P),\\
 \end{aligned}
\end{equation}
The appearance $\bm z$ can also be encoded from an input image by using the encoder on its partial texture map  $\bm T$, \textit{i.e.,} $\bm z = \bm \mu$, where $\bm \mu, \bm \sigma = E(\bm T)$. Keeping $\bm z$ fixed and varying the the pose $\bm P$, we can perform \textit{pose transfer} \cite{Sarkar2020,Grigorev2019CoordinateBasedTI,Siarohin2019AppearanceAP,Neverova2018}, \textit{i.e.,} re-rendering a subject with  different pose and viewpoint. 
We can also perform \textit{parts-based sampling} and \textit{garment transfer} by only varying the vector $\bm z[k]$ corresponding to the part $k$.
See Sec.~\ref{sec:results} for all possible  applications of our system. 
%


\section{Experimental Results} 
\label{sec:results}

\noindent\textbf{Dataset.} 
We use the \textit{In-shop Clothes Retrieval Benchmark} of DeepFashion dataset \cite{Liu2016DeepFashion} for our main experiments. 
The dataset comprises of around 52K high-resolution images of fashion models with 13K different clothing items in different poses. Training and testing splits are also provided. To filter non-human images, we discard all the images where we could not compute DensePose, resulting in 38K training images and 3K testing images. We train our system with the resulting training split and use the testing split for conditioning poses. We also show qualitative results of our method with Fashion dataset \cite{Zablotskaia2019DwNetDW} that has 500 training and 100 test videos, each containing roughly 350 frames.

\noindent\textbf{Experimental Setup.} 
We train our model for the resolution of 512 $\times$ 512 with the training procedure described in Sec.~\ref{sec:training}. The resolution of the partial texture map $T$ is chosen to be 256 $\times$ 256, number of body parts $M = 24$, and the latent vector dimension $N = 16$. To evaluate our model for the ability to generate diverse and realistic images, preserve the identity of the generated output across different poses, and perform part-based sampling, we use the \textit{same trained model} for \textit{all the experiments} in the following subsections. 
All the ablation experiments use the same setting in terms of output resolution. When the result of the comparison methods do not have a trained model for 512 resolution (especially for the SOA methods on pose transfer), we resize our image before performing the quantitative evaluation.

\begin{figure*}[t]
    \includegraphics[width=\linewidth]{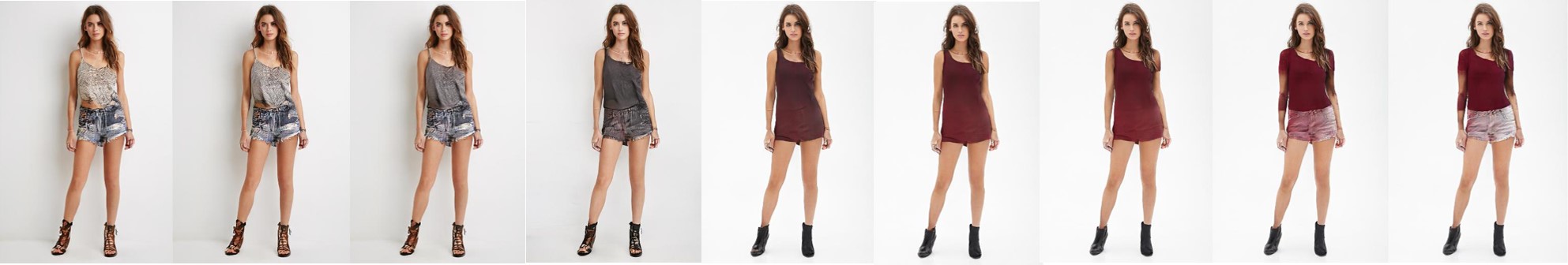}
    \caption{Generated images with interpolated appearance encodings. Note that each of the intermediate images is a coherently dressed human image. 
    } 
    \label{fig:latent_inter} 
\end{figure*}
\subsection{Appearance Sampling} 
\label{sec:res:appearance}
We next evaluate the ability of our system to generate diverse and realistic images of humans given a conditioning pose. Given a \textit{fixed pose}, we \textit{randomly generate samples} from the latent vector $\bm z \sim \mathcal{N}(0^{MN}, I^{MN})$ and compare our method with the following baseline methods. 

\begin{table}[t]
\small
\centering
\begin{tabular}{c|c|c} 
              & \textbf{Diversity}    & \textbf{Realism}   \\ 
              &LPIPS Distance $\uparrow$& FID $\downarrow$\\\hline
VUnet  \cite{esser2018variational}           &   0.182      &     50.0  \\              
Pix2PixHD+Noise          &  4.6e-6    &   109.4    \\
Pix2PixHD+WNoise          &  0.008        &   101.9    \\
DAE+WNoise        &  0.083      &   187.4   \\
\textbf{Ours} & \textbf{0.219}  &  \textbf{24.9}   \\ \hline
ground truth  & 0.44 &  0.0  \\ 
\end{tabular}
\caption{Diversity \textit{vs} realism. We use LPIPS distance \cite{zhang2018perceptual} between the randomly generated samples of the same pose to measure Diversity, and FID to measure the realism of the generated samples. $\uparrow$ ($\downarrow$) means higher (lower) is better.} 
\label{table:appearance}
\vspace{-0.3cm}
\end{table}

\noindent\textbf{Pix2PixHD + Noise.} 
Some generators, such as Pix2Pix, Pix2PixHD produces a single output given a conditional input. 
Randomly drawn noise from a prior distribution can be added to the input of conditional generators to induce stochasticity. We sample noise $z \in \mathbb{R}^N$ from the standard Gaussian distribution and tile it across the 3 channel condition DensePose image to produce $3+N$ channel input vector and train Pix2PixHD for a pose-conditioned multimodal human generator. 
We optimize the conditional generator $G$ and discriminator $D$ with the GAN loss:
$\max\limits_{D}\min\limits_{G} \mathbb{E}_{\bm{P,T} \sim p(\bm{P,T)}}[\log(D(\bm{P,I}))] + \mathbb{E}_{\bm P\sim p(\bm P), \bm z \sim p(\bm z)}[\log(1-D(\bm P, G(\bm P, \bm z))]$, and the reconstruction loss between $G(\bm P, \bm z)$ and $\bm I$.

\noindent\textbf{Pix2PixHD + Warped Noise.} 
Isola \textit{et al.}~\cite{pix2pix2017} and Zhu \textit{et al.}~\cite{NIPS2017_6650} observed that a simple extension of a translation network with noise for the purpose multimodal generation often ends up in mode collapse. Therefore, we extend our \textit{Pix2PixHD + Noise} baseline by sampling part specific noise vector $\bm z \in \mathbb{R}^{M \times N}$ and warping it with the input DensePose to create a noise image $\bm Z_P$ as in Sec.~\ref{sec:warping}. This noise image is used as conditioning input to Pix2PixHD. 

\noindent\textbf{Deterministic Auto Encoder with Warped Noise (DAE).} 
This baseline serves as the deterministic version of our method. Because of the lack of constraints on the latent variable (KL divergence with the prior), and direct broadcasting of a single noise vector to multiple pixels in the warping operation, we encountered the problem of exploding gradients. To make the training process tractable, we use a UNet to encode the appearance to the UV texture space, $\bm Z_s = E(\bm T_s), \bm Z_s \in \mathbb{R}^{h \times w \times N}$ ($h, w$ are the texture map dimensions). 
We warp $\bm Z_s$ from the texture coordinates to the pixel coordinates by the target pose (instead of a single vector per parts) to complete the pipeline for training. 
Because of the similarity of this baseline to NHRR~\cite{Sarkar2020}, we use UNet configuration as in their work. 

\begin{figure}[t]
    \includegraphics[width=\linewidth]{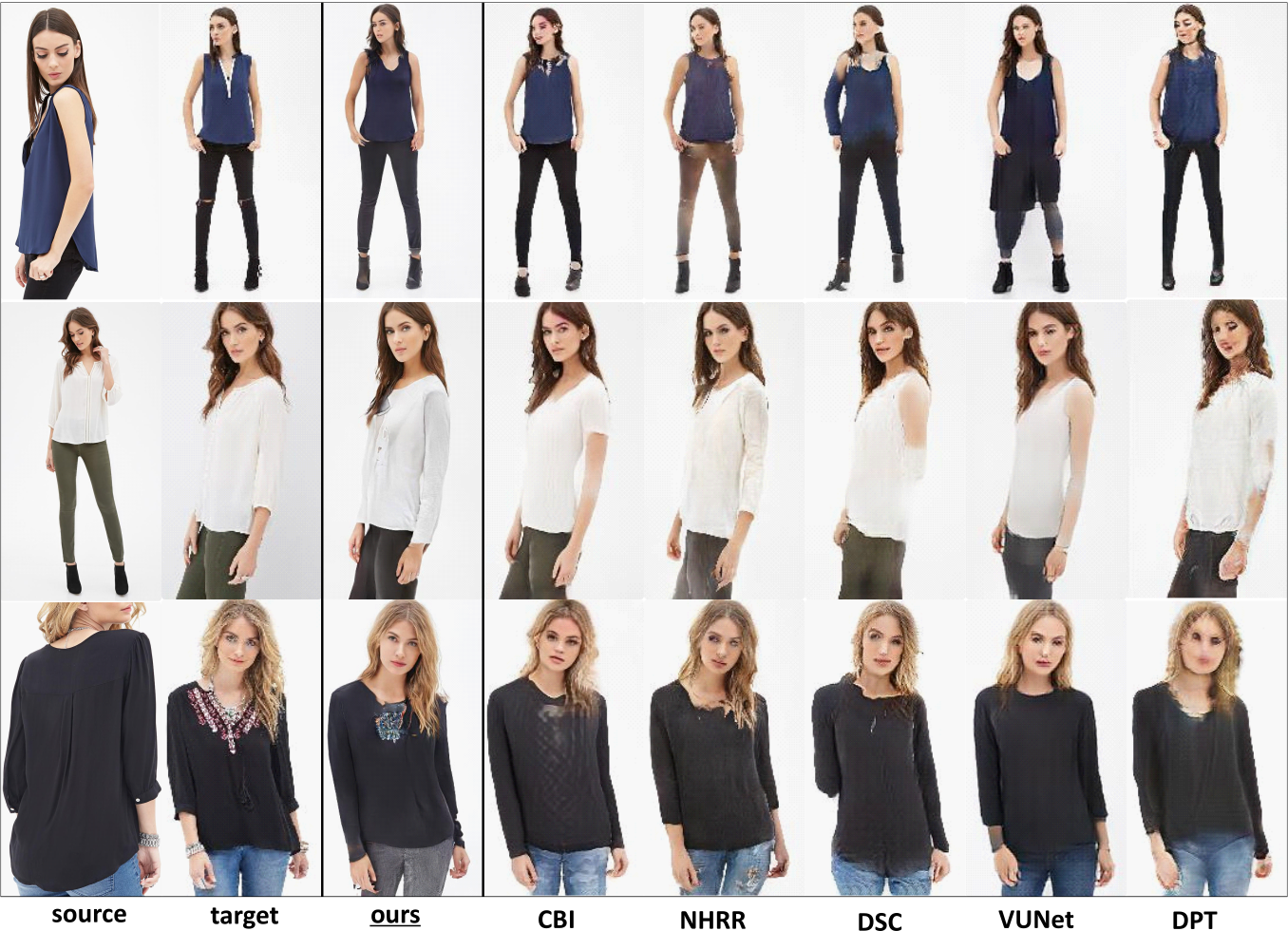}
    \caption{
    Comparison of our reconstruction+transfer results with the state-of-the-art pose transfer methods: CBI \cite{Grigorev2019CoordinateBasedTI},  NHRR \cite{Sarkar2020},  DSC \cite{Siarohin2019AppearanceAP}, VUNet \cite{esser2018variational}, and DPT. Our approach produces more realistic renderings than the competing methods. Best viewed when zoomed digitally. More results are provided in the appendix.} 
    \label{fig:transfer} 
    \vspace{-0.3cm}
\end{figure}

\begin{table}[t] 
\resizebox{\columnwidth}{!}{%
\centering 
\begin{tabular}{@{}c|c|c||c|c|c@{}} 
              & SSIM $\uparrow$   & LPIPS  $\downarrow$ & & SSIM $\uparrow$   & LPIPS  $\downarrow$ \\ \hline
CBI  \cite{Grigorev2019CoordinateBasedTI}          &  0.766     &   0.178 &DPT \cite{Neverova2018}           &   0.759      &     0.206     \\
DSC \cite{Siarohin2019AppearanceAP}          &  0.750        &   0.214     & NHRR \cite{Sarkar2020} & {0.768}  &  \textbf{0.164}   \\ 
VUNet  \cite{esser2018variational}       &  0.739      &    0.202  & \textbf{Ours} & \textbf{0.777}  &  {0.187}  \\ \hline
\end{tabular}} 
\caption{Quantitative results for pose transfer against several  state-of-the-art methods using Structural Similarity Index (SSIM) \cite{ssim2004} and Learned Perceptual Image Patch Similarity (LPIPS)  \cite{zhang2018perceptual}. $\uparrow$ ($\downarrow$) means higher (lower) is better.} 
\label{table:transfer} 
\vspace{-0.4cm} 
\end{table} 

\noindent\textbf{VUNet \cite{esser2018variational}.} 
We compare our result to VUNet that performs disentanglement between appearance and structure and can be used for human generation. We use their publically available code and trained model on DeepFashion dataset and report their results here. Note that in contrast to our  method, VUNet cannot perform part-based sampling  (Sec.~\ref{sec:res:parts}). 

\noindent\textbf{Additional Baselines} We perform following additional baselines during development and training of our model: a) \textit{NoParts} does not perform part-specific warping, but broadcasts concatenated part-encoded noise-vector to the human silhouette (see Sec. \ref{sec:res:parts} for more details on this) 
b) \textit{+DPCond} conditions the DensePose image $\bm P_t$ in addition to the noise image $\bm Z_t$ to the generator. c) \textit{+NoisePrior} use samples from the prior $\mathcal{N}(0,1)$  (along with the encoded distribution) during the training, as recommended by Larsen \textit{et al.}~\cite{larsen2016autoencoding}. All the aforementioned \textit{additional baselines} performed poorly, and we present their results in the appendix. We discuss here the main baselines that are most representative of the different methods.
\begin{figure*}[t]
    \includegraphics[width=\linewidth]{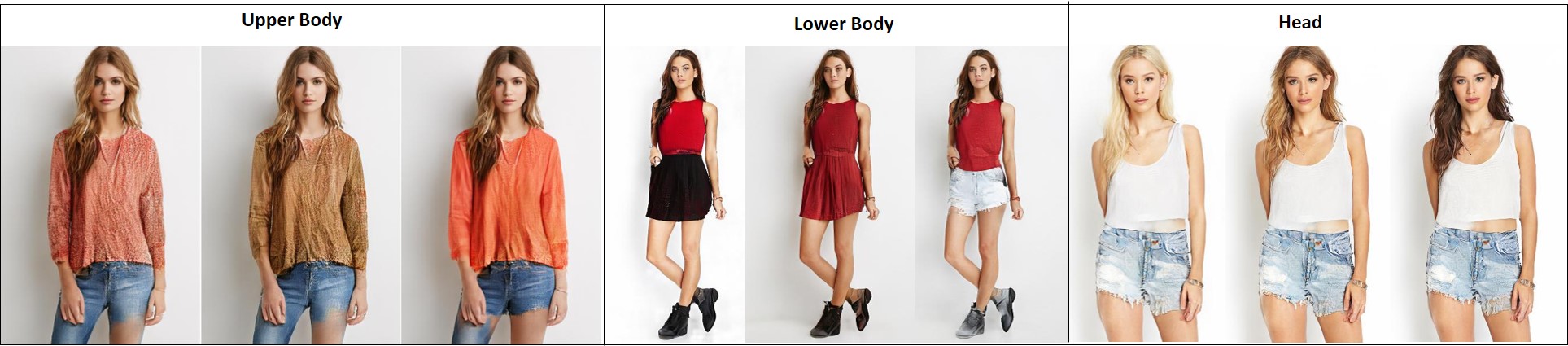}
    \caption{Our results for part sampling. 
    We change the latent noise corresponding to a specific body part. 
    %
    } 
    \label{fig:parts} 
\end{figure*}
\begin{figure}[t]
    \includegraphics[width=\linewidth]{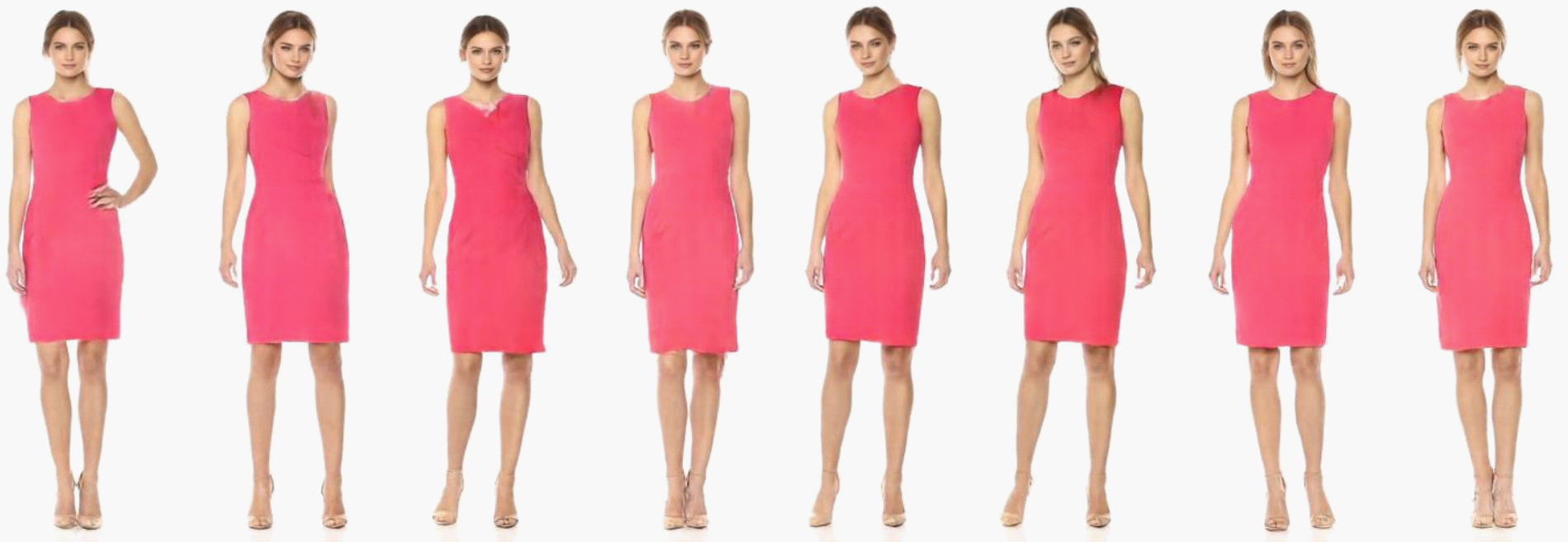} 
    \caption{Results of our method for motion transfer in a walking  sequence from fashion dataset \cite{Zablotskaia2019DwNetDW}. See the accompanying video for the motion results.
    } 
    \label{fig:fashion} 
    \vspace{-0.5cm}
\end{figure}

The qualitative results are shown in Fig.~\ref{fig:appearance}. 
We find that \textit{Pix2PixHD+N} produces a single realistic output on the conditioning pose and undergo full mode collapse. The baseline \textit{Pix2PixHD+WN} produces slightly diverse output than \textit{Pix2PixHD+N}, but the variations are still not meaningful. While the deterministic method of \textit{DAE} is a right choice for reconstruction and pose transfer (see results of NHRR \cite{Sarkar2020} in Sec.~\ref{sec:res:posetransfer}), 
there is no guarantee that the distribution of the encoded latent space will be close to a prior distribution, which makes the test-time sampling difficult. This makes the output far from realistic when the latent vector is sampled from  $\mathcal{N}(0, I)$. VUNet produces a diverse output  but lacks the quality due to its less powerful generator. In contrast, we find that our full method produces results that are both diverse and realistic. We also find our latent space of appearance to be smooth and interpolateable. Fig.~\ref{fig:latent_inter} shows images generated by interpolating the appearance vector $\bm z$ between two encodings. Note how each of the intermediate images is a coherently dressed human image. More qualitative results are provided in the appendix. 
To evaluate our method quantitatively, we randomly select 100 poses from the test set and generate 50 samples for each pose for all the methods. We measure diversity and realism of all the baselines by the following metrics \textit{1) Pairwise LPIPS distance} -- we compute LPIPS distance \cite{zhang2018perceptual} between all the generated samples for each pose, and take the mean of the distances of all such pairs. More the value of this metric, more diverse is the output. \textit{2) FID}-- we compute the Frech{\^e}t Inception Distance between the generated samples and the training split of the dataset. FID captures how close is the distribution of the generated samples, from the distribution of the ground truth in the InceptionV3 feature space, and it has been used widely in the community as a metric for quality for GANs  \cite{Brock2018Biggan,karras2019style,tewari2020stylerig,tewari2020pie}. 
The quantitative results are in Table \ref{table:appearance}. Our method outperforms other baselines significantly in terms of quality of the image (FID), while maintaining diversity. 

\noindent
\textbf{User Study.}
We perform a comprehensive user study to access visual fidelity and characteristics of the results between VUnet and our method for appearance sampling, as those arguably generate visually most realistic human images (see Table \ref{table:appearance}). To this end, we use $20$ random test poses and choose the results by both the methods which look most realistic, in our opinion. %
Users are then shown a pair of generated images from the two methods for 12 poses and asked to select the most realistic one among them. Since such comparisons on individual images can be biased, we also ask the participants to decide between image sets (two to four images) for eight poses. The purpose of those is to compensate for the possible image selection biases associated with image pairs. 
In the total of $36$ respondents, our method is preferred over VUnet in $\bf 91.06\%$ of the cases.

\subsection{Pose Transfer} 
\label{sec:res:posetransfer} 
In this section, we evaluate how our system preserves the appearance of the outputs across different poses and  perform the following pose transfer experiment, \textit{i.e.,} re-rendering of a subject under different poses and viewpoints. 
We encode the appearance $\bm T$ of the input image $\bm I$ by using the mean of the distribution predicted by the encoder, \textit{i.e.,} $\bm z = \bm \mu$ where $\bm \mu, \bm \sigma = E(\bm T)$. 
We keep $\bm z$ \textit{fixed}, and use \textit{different target poses} to generate images for pose transfer. 

\begin{figure*}[t]
    \includegraphics[width=\linewidth]{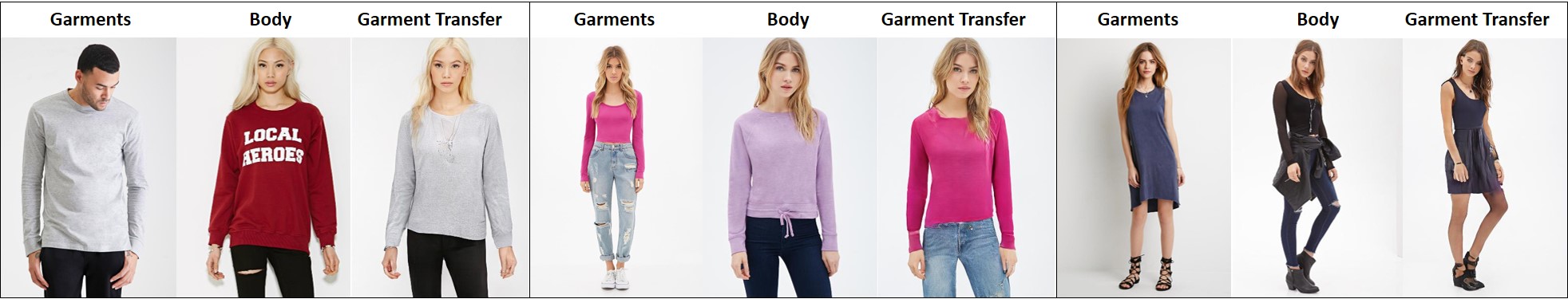}
    \caption{The results of our method for garment transfer. By combining the appearance encoding of two images based on the body parts, we can perform garment transfer.} 
    \label{fig:garments} 
    \vspace{-0.3cm}
\end{figure*} 

We compare our results with five state-of-the-art pose transfer methods, namely Coordinate Based Inpainting (CBI) \cite{Grigorev2019CoordinateBasedTI}, Deformable GAN (DSC) \cite{Siarohin2019AppearanceAP}, Variational U-Net (VUNet) \cite{esser2018variational} Dense Pose Transfer (DPT) \cite{Neverova2018} and Neural Human Re-Rendering (NHRR) \cite{Sarkar2020}. For both qualitative and quantitative evaluation, we use the results of 176 testing pairs that are used in the existing work \cite{Sarkar2020,Grigorev2019CoordinateBasedTI}. The qualitative results are shown in Fig.~\ref{fig:transfer}. Note that these methods, except for VUNet, are designed explicitly for the problem of pose transfer, while our method is designed as a generative model which is capable of retaining identity across different pose. We observe that our results show better realism than the other state-of-the-art methods, and perform comparably for the problem of pose transfer. This is confirmed by the comparable LPIPS distance of our model in comparison to the other SOA methods (Table \ref{table:transfer}). The appearance consistency is also verified by our result on the walking sequences of the fashion dataset \cite{Zablotskaia2019DwNetDW} (Fig.~\ref{fig:fashion}).

\noindent
\textbf{User Study.}
Following the existing work on pose transfer, we perform another user study for evaluating our results for this task, and compare with the method of NHRR and CBI.
We adapt the user study format and question types from 
\cite{Sarkar2020} and show the generated results of the methods from 16 source-target pairs. The user preference for \textit{identity preservation} was found to be \textbf{ours: 65.62\%}, CBI: 21.88\%, NHRR: 12.5\%. The user preference for \textit{realism} was found to be \textbf{ours: 81.25\%}, CBI: 6.25\%,  NHRR: 12.5\%.

\begin{table}[t]
\small
\centering
\begin{tabular}{c|c|c} 
              & Variation--Part $\uparrow$   & Variation--Rest  $\downarrow$ \\ \hline
NoParts          &  0.45     &  0.44    \\
\textbf{Ours} & 0.37  &  0.11\\    \hline
\end{tabular}
\caption{Quantitative evaluation for part sampling using mean pairwise L1 distance of the masked part. $\downarrow$ ($\uparrow$) means lower (higher) is better.}
\label{table:parts}
\vspace{-0.5cm}
\end{table}

\subsection{Part-Based Sampling} 
\label{sec:res:parts} 

We next evaluate our method for part-based sampling -- the ability to produce different plausible renderings of a body part (\textit{e.g.,} head) while keeping the rest of the body same. 
To this end, we vary the vector $\bm z[k] \sim \mathcal{N}(0^{N}, I^{N}) $ corresponding to the part $k$, and keep the rest of the noise vector $\bm z[j] \mid j \neq k$ fixed, and perform the decoding on a given pose. When multiple elementary DensePose parts (\textit{e.g.,} left head, right head) correspond to one logical body part for sampling (\textit{e.g.,} head), we sample noise in all the elementary part vectors. The results are shown in Fig.~\ref{fig:parts}.

To explicitly see how much our design choices help for part sampling, we perform a baseline experiment \textbf{NoParts} where we encode the appearance in a single vector $\bm z \sim E(\bm T) \mid \bm z \in \mathbb{Z}^{N}$ instead of part-specific vectors. We then warp this vector using the conditioning DensePose to create a noise image for the generator as described in Sec.~\ref{sec:warping}. 
We compute the following two metrics for a given part $p$: 1) \textit{Variation--Part}: mean pairwise L1 distance between the samples in the masked region (by DensePose) of the part $p$ normalized by the masked area. 2) \textit{Variation--Rest}: mean pairwise L1 distance between the samples in the masked region of all body parts excluding $p$. We compute the aforementioned metrics for the following parts: ``Head'', ``Upper body'' and ``Lower body'' for 2500 generated images and provide our result in Table \ref{table:parts}. A suitable method for part sampling should generate diverse semantically meaningful renderings of a part without changing  the rest of the body. This is confirmed by high  \textit{Variation--Part} and low \textit{Variation--Rest} in our full method in comparison to the baseline \textit{NoParts}. 

Using part-specific latent vectors allows us to naturally perform

\section{Limitations and Future Work}
Our method sometimes demonstrates spurious interleaving of body parts or garments in the generated images (\textit{e.g.,} see  Fig.~\ref{fig:latent_inter}, right-most images). 
However, this is also shared by other (less realistic) generative human models (see the result of VUNet in  Fig.~\ref{fig:appearance} and the baselines in Fig.~\ref{fig:transfer}). These artifacts could be avoided by a hierarchical generator, where the garment style is generated first. This design, however, comes with the disadvantage of not being able to perform part-based sampling. We have also found that our generated images are biased towards females. This, we hypothesize, is due to the bias in the DeepFashion dataset.

\textit{garment transfer} between two images representing the body and garments. To this end, we first encode the appearance of both the body image $\bm I_b$ and garment image $\bm I_g$, in their part-based noise vectors $\bm z_b$ and $\bm z_g$ receptively. We then construct a new noise embedding that comprises of the body parts $\bm z_b[p] \mid p \in$ \textit{Body}, and garment parts $\bm z_g[p] \mid p \in$ \textit{Garments} of the two noise embeddings, and use it in the generator for the final output, see
Fig.~\ref{fig:garments}. 

\section{Conclusion} 
\label{sec:conclusion} 
We  have presented a generative model for full-body images of clothed humans, which enables control of body pose, as well as independent control and sampling of appearance and clothing style on a body part level. 
A framework based on variational autoencoders is used to induce stochasticity in the appearance space. 
To achieve the disentanglement of pose and appearance, we encode the posterior probability of the part-specific latent vectors from a space of pose-independent appearance and warp the encoded vector to a different pose before performing the reconstruction. 
Experiments with pose-conditioned image generation, pose transfer, as well as parts and garment transfer, have demonstrated that the model improves over the state-of-the-arts. 

\noindent\textbf{Acknowledgements.} This work was supported by the ERC Consolidator Grant 4DReply (770784) and Lise Meitner Postdoctoral Fellowship. 
We thank Dushyant Mehta for the feedback on the draft. 

{\small
\bibliographystyle{ieee_fullname}
\bibliography{article}

\begin{thebibliography}{10}\itemsep=-1pt

\bibitem{Lischinski2018}
Kfir Aberman, Mingyi Shi, Jing Liao, Dani Lischinski, Baoquan Chen, and Daniel
  Cohen{-}Or.
\newblock Deep video-based performance cloning.
\newblock {\em Comput. Graph. Forum}, 38(2):219--233, 2019.

\bibitem{Recycle-GAN}
Aayush Bansal, Shugao Ma, Deva Ramanan, and Yaser Sheikh.
\newblock Recycle-gan: Unsupervised video retargeting.
\newblock In {\em ECCV}, 2018.

\bibitem{Brock2018Biggan}
Andrew Brock, Jeff Donahue, and Karen Simonyan.
\newblock Large scale {GAN} training for high fidelity natural image synthesis.
\newblock {\em CoRR}, abs/1809.11096, 2018.

\bibitem{chan2019dance}
Caroline Chan, Shiry Ginosar, Tinghui Zhou, and Alexei~A Efros.
\newblock Everybody dance now.
\newblock In {\em International Conference on Computer Vision (ICCV)}, 2019.

\bibitem{choi2017stargan}
Yunjey Choi, Minje Choi, Munyoung Kim, Jung-Woo Ha, Sunghun Kim, and Jaegul
  Choo.
\newblock {StarGAN}: Unified generative adversarial networks for multi-domain
  image-to-image translation.
\newblock In {\em Computer Vision and Pattern Recognition (CVPR), 2018}, 2018.

\bibitem{Doan2019OnlineAC}
Thang Doan, J. Monteiro, Isabela Albuquerque, Bogdan Mazoure, A. Durand, Joelle
  Pineau, and R.~Devon Hjelm.
\newblock Online adaptative curriculum learning for gans.
\newblock In {\em AAAI}, 2019.

\bibitem{GMAN}
Ishan Durugkar, Ian Gemp, and Sridhar Mahadevan.
\newblock Generative multi-adversarial networks, 2017.

\bibitem{esser2018variational}
Patrick Esser, Ekaterina Sutter, and Bj{\"o}rn Ommer.
\newblock A variational u-net for conditional appearance and shape generation.
\newblock In {\em Computer Vision and Pattern Recognition (CVPR)}, pages
  8857--8866, 2018.

\bibitem{Goodfellow14}
Ian Goodfellow, Jean Pouget-Abadie, Mehdi Mirza, Bing Xu, David Warde-Farley,
  Sherjil Ozair, Aaron Courville, and Yoshua Bengio.
\newblock Generative adversarial nets.
\newblock In Z. Ghahramani, M. Welling, C. Cortes, N. Lawrence, and K.~Q.
  Weinberger, editors, {\em Advances in Neural Information Processing Systems},
  volume~27, pages 2672--2680. Curran Associates, Inc., 2014.

\bibitem{Grigorev2019CoordinateBasedTI}
A.~K. Grigor'ev, Artem Sevastopolsky, Alexander Vakhitov, and Victor~S.
  Lempitsky.
\newblock Coordinate-based texture inpainting for pose-guided human image
  generation.
\newblock {\em Computer Vision and Pattern Recognition (CVPR)}, pages
  12127--12136, 2019.

\bibitem{pix2pix2017}
Phillip Isola, Jun-Yan Zhu, Tinghui Zhou, and Alexei~A Efros.
\newblock Image-to-image translation with conditional adversarial networks.
\newblock {\em CVPR}, 2017.

\bibitem{karras2018progressive}
Tero Karras, Timo Aila, Samuli Laine, and Jaakko Lehtinen.
\newblock Progressive growing of {GAN}s for improved quality, stability, and
  variation.
\newblock In {\em International Conference on Learning Representations}, 2018.

\bibitem{karras2019style}
Tero Karras, Samuli Laine, and Timo Aila.
\newblock A style-based generator architecture for generative adversarial
  networks.
\newblock In {\em Proceedings of the IEEE conference on computer vision and
  pattern recognition}, pages 4401--4410, 2019.

\bibitem{Karras2019stylegan2}
Tero Karras, Samuli Laine, Miika Aittala, Janne Hellsten, Jaakko Lehtinen, and
  Timo Aila.
\newblock Analyzing and improving the image quality of {StyleGAN}.
\newblock In {\em Proc. CVPR}, 2020.

\bibitem{karras2020analyzing}
Tero Karras, Samuli Laine, Miika Aittala, Janne Hellsten, Jaakko Lehtinen, and
  Timo Aila.
\newblock Analyzing and improving the image quality of stylegan.
\newblock In {\em Proceedings of the IEEE/CVF Conference on Computer Vision and
  Pattern Recognition}, pages 8110--8119, 2020.

\bibitem{adam}
Diederick~P Kingma and Jimmy Ba.
\newblock Adam: A method for stochastic optimization.
\newblock In {\em International Conference on Learning Representations (ICLR)},
  2015.

\bibitem{Kingma2014VAE}
Diederik~P. Kingma and Max Welling.
\newblock Auto-encoding variational bayes.
\newblock In Yoshua Bengio and Yann LeCun, editors, {\em 2nd International
  Conference on Learning Representations, {ICLR} 2014, Banff, AB, Canada, April
  14-16, 2014, Conference Track Proceedings}, 2014.

\bibitem{KratzHPV2017}
Bernhard Kratzwald, Zhiwu Huang, Danda~Pani Paudel, and Luc Van~Gool.
\newblock Towards an understanding of our world by {GANing} videos in the wild.
\newblock arXiv:1711.11453, 2017.

\bibitem{larsen2016autoencoding}
Anders Boesen~Lindbo Larsen, S{\o}ren~Kaae S{\o}nderby, Hugo Larochelle, and
  Ole Winther.
\newblock Autoencoding beyond pixels using a learned similarity metric.
\newblock In {\em International conference on machine learning}, pages
  1558--1566. PMLR, 2016.

\bibitem{Lassner2017}
Christoph Lassner, Gerard Pons-Moll, and Peter~V. Gehler.
\newblock A generative model for people in clothing.
\newblock In {\em Proceedings of the IEEE International Conference on Computer
  Vision}, 2017.

\bibitem{liu2020NeuralHumanRendering}
Lingjie Liu, Weipeng Xu, Marc Habermann, Michael Zollhöfer, Florian Bernard,
  Hyeongwoo Kim, Wenping Wang, and Christian Theobalt.
\newblock Neural human video rendering by learning dynamic textures and
  rendering-to-video translation.
\newblock {\em IEEE Transactions on Visualization and Computer Graphics},
  PP:1--1, 05 2020.

\bibitem{Liu2019}
Lingjie Liu, Weipeng Xu, Michael Zollhoefer, Hyeongwoo Kim, Florian Bernard,
  Marc Habermann, Wenping Wang, and Christian Theobalt.
\newblock Neural rendering and reenactment of human actor videos.
\newblock {\em ACM Transactions on Graphics (TOG)}, 2019.

\bibitem{LiuBK2017}
Ming-Yu Liu, Thomas Breuel, and Jan Kautz.
\newblock Unsupervised image-to-image translation networks.
\newblock In {\em Proceedings of the 31st International Conference on Neural
  Information Processing Systems}, NIPS’17, pages 700–--708, Red Hook, NY,
  USA, 2017. Curran Associates Inc.

\bibitem{liu2017sphereface}
Weiyang Liu, Yandong Wen, Zhiding Yu, Ming Li, Bhiksha Raj, and Le Song.
\newblock Sphereface: Deep hypersphere embedding for face recognition.
\newblock In {\em Computer Vision and Pattern Recognition (CVPR)}, pages
  212--220, 2017.

\bibitem{Liu2016DeepFashion}
Z. {Liu}, P. {Luo}, S. {Qiu}, X. {Wang}, and X. {Tang}.
\newblock Deepfashion: Powering robust clothes recognition and retrieval with
  rich annotations.
\newblock In {\em Computer Vision and Pattern Recognition (CVPR)}, pages
  1096--1104, 2016.

\bibitem{SMPL:2015}
Matthew Loper, Naureen Mahmood, Javier Romero, Gerard Pons-Moll, and Michael~J.
  Black.
\newblock {SMPL}: A skinned multi-person linear model.
\newblock {\em ACM Trans. Graphics (Proc. SIGGRAPH Asia)}, 34(6):248:1--248:16,
  Oct. 2015.

\bibitem{MaSJSTV2017}
Liqian Ma, Xu Jia, Qianru Sun, Bernt Schiele, Tinne Tuytelaars, and Luc
  Van~Gool.
\newblock Pose guided person image generation.
\newblock In {\em Advances in Neural Information Processing Systems}, pages
  405--415, 2017.

\bibitem{Ma18}
Liqian Ma, Qianru Sun, Stamatios Georgoulis, Luc van Gool, Bernt Schiele, and
  Mario Fritz.
\newblock Disentangled person image generation.
\newblock {\em Computer Vision and Pattern Recognition (CVPR)}, 2018.

\bibitem{mirza2014conditional}
Mehdi Mirza and Simon Osindero.
\newblock Conditional generative adversarial nets, 2014.

\bibitem{miyato2018cgans}
Takeru Miyato and Masanori Koyama.
\newblock c{GAN}s with projection discriminator.
\newblock In {\em International Conference on Learning Representations}, 2018.

\bibitem{Mordido2018DropoutGANLF}
Gonçalo Mordido, Haojin Yang, and C. Meinel.
\newblock Dropout-gan: Learning from a dynamic ensemble of discriminators.
\newblock {\em ArXiv}, abs/1807.11346, 2018.

\bibitem{Neverova2018}
Natalia Neverova, Riza~Alp G\"{u}ler, and Iasonas Kokkinos.
\newblock Dense pose transfer.
\newblock {\em European Conference on Computer Vision (ECCV)}, 2018.

\bibitem{pmlr-v70-odena17a}
Augustus Odena, Christopher Olah, and Jonathon Shlens.
\newblock Conditional image synthesis with auxiliary classifier {GAN}s.
\newblock volume~70 of {\em Proceedings of Machine Learning Research}, pages
  2642--2651, International Convention Centre, Sydney, Australia, 06--11 Aug
  2017. PMLR.

\bibitem{park2019SPADE}
Taesung Park, Ming-Yu Liu, Ting-Chun Wang, and Jun-Yan Zhu.
\newblock Semantic image synthesis with spatially-adaptive normalization.
\newblock In {\em Proceedings of the IEEE Conference on Computer Vision and
  Pattern Recognition}, 2019.

\bibitem{Pumarola_2018_CVPR}
Albert Pumarola, Antonio Agudo, Alberto Sanfeliu, and Francesc Moreno-Noguer.
\newblock Unsupervised person image synthesis in arbitrary poses.
\newblock In {\em The IEEE Conference on Computer Vision and Pattern
  Recognition (CVPR)}, June 2018.

\bibitem{Guler2018DensePose}
Iasonas~Kokkinos Rieza Alp~Gueler, Natalia~Neverova.
\newblock Densepose: Dense human pose estimation in the wild.
\newblock In {\em Computer Vision and Pattern Recognition (CVPR)}, 2018.

\bibitem{Sarkar2020}
Kripasindhu Sarkar, Dushyant Mehta, Weipeng Xu, Vladislav Golyanik, and
  Christian Theobalt.
\newblock Neural re-rendering of humans from a single image.
\newblock In {\em European Conference on Computer Vision (ECCV)}, 2020.

\bibitem{Siarohin2019AppearanceAP}
Aliaksandr Siarohin, St{\'e}phane Lathuili{\`e}re, Enver Sangineto, and Nicu
  Sebe.
\newblock Appearance and pose-conditioned human image generation using
  deformable gans.
\newblock {\em Transactions on Pattern Analysis and Machine Intelligence
  (TPAMI)}, 2019.

\bibitem{SiaroSLS2017}
Aliaksandr Siarohin, Enver Sangineto, Stephane Lathuiliere, and Nicu Sebe.
\newblock Deformable {GANs} for pose-based human image generation.
\newblock In {\em CVPR 2018}, 2018.

\bibitem{simonyan2014very}
Karen Simonyan and Andrew Zisserman.
\newblock Very deep convolutional networks for large-scale image recognition.
\newblock {\em arXiv preprint arXiv:1409.1556}, 2014.

\bibitem{tewari2020stylerig}
Ayush Tewari, Mohamed Elgharib, Gaurav Bharaj, Florian Bernard, Hans-Peter
  Seidel, Patrick P{\'e}rez, Michael Z{\"o}llhofer, and Christian Theobalt.
\newblock Stylerig: Rigging stylegan for 3d control over portrait images, cvpr
  2020.
\newblock In {\em {IEEE} Conference on Computer Vision and Pattern Recognition
  (CVPR)}. {IEEE}, june 2020.

\bibitem{tewari2020pie}
Ayush Tewari, Mohamed Elgharib, Mallikarjun BR, Florian Bernard, Hans-Peter
  Seidel, Patrick P{\'e}rez, Michael Z{\"o}llhofer, and Christian Theobalt.
\newblock Pie: Portrait image embedding for semantic control.
\newblock volume~39, December 2020.

\bibitem{varol17_surreal}
G{\"u}l Varol, Javier Romero, Xavier Martin, Naureen Mahmood, Michael~J. Black,
  Ivan Laptev, and Cordelia Schmid.
\newblock Learning from synthetic humans.
\newblock In {\em Computer Vision and Pattern Regognition (CVPR)}, 2017.

\bibitem{wang2018vid2vid}
Ting-Chun Wang, Ming-Yu Liu, Jun-Yan Zhu, Guilin Liu, Andrew Tao, Jan Kautz,
  and Bryan Catanzaro.
\newblock Video-to-video synthesis.
\newblock In {\em Advances in Neural Information Processing Systems (NeurIPS)},
  2018.

\bibitem{wang2018pix2pixHD}
Ting-Chun Wang, Ming-Yu Liu, Jun-Yan Zhu, Andrew Tao, Jan Kautz, and Bryan
  Catanzaro.
\newblock High-resolution image synthesis and semantic manipulation with
  conditional gans.
\newblock In {\em Computer Vision and Pattern Recognition (CVPR)}, 2018.

\bibitem{YiZTG2017}
Zili Yi, Hao Zhang, Ping Tan, and Minglun Gong.
\newblock {DualGAN}: Unsupervised dual learning for image-to-image translation.
\newblock In {\em ICCV}, pages 2868--2876, Oct. 2017.

\bibitem{Zablotskaia2019DwNetDW}
Polina Zablotskaia, Aliaksandr Siarohin, Leonid Sigal, and Bo Zhao.
\newblock Dwnet: Dense warp-based network for pose-guided human video
  generation.
\newblock In {\em British Machine Vision Conference (BMVC)}, 2019.

\bibitem{sagan}
Han Zhang, Ian Goodfellow, Dimitris Metaxas, and Augustus Odena.
\newblock Self-attention generative adversarial networks.
\newblock volume~97 of {\em Proceedings of Machine Learning Research}, pages
  7354--7363, Long Beach, California, USA, 09--15 Jun 2019. PMLR.

\bibitem{zhang2018perceptual}
Richard Zhang, Phillip Isola, Alexei~A Efros, Eli Shechtman, and Oliver Wang.
\newblock The unreasonable effectiveness of deep features as a perceptual
  metric.
\newblock In {\em Computer Vision and Pattern Recognition (CVPR)}, 2018.

\bibitem{ssim2004}
{Zhou Wang}, A.~C. {Bovik}, H.~R. {Sheikh}, and E.~P. {Simoncelli}.
\newblock Image quality assessment: from error visibility to structural
  similarity.
\newblock {\em IEEE Transactions on Image Processing}, 13(4):600--612, 2004.

\bibitem{CycleGAN2017}
Jun-Yan Zhu, Taesung Park, Phillip Isola, and Alexei~A Efros.
\newblock Unpaired image-to-image translation using cycle-consistent
  adversarial networkss.
\newblock In {\em Computer Vision (ICCV), 2017 IEEE International Conference
  on}, 2017.

\bibitem{NIPS2017_6650}
Jun-Yan Zhu, Richard Zhang, Deepak Pathak, Trevor Darrell, Alexei~A Efros,
  Oliver Wang, and Eli Shechtman.
\newblock Toward multimodal image-to-image translation.
\newblock In I. Guyon, U.~V. Luxburg, S. Bengio, H. Wallach, R. Fergus, S.
  Vishwanathan, and R. Garnett, editors, {\em Advances in Neural Information
  Processing Systems 30}, pages 465--476. Curran Associates, Inc., 2017.

\bibitem{zhu2019progressive}
Zhen Zhu, Tengteng Huang, Baoguang Shi, Miao Yu, Bofei Wang, and Xiang Bai.
\newblock Progressive pose attention transfer for person image generation.
\newblock In {\em Proceedings of the IEEE Conference on Computer Vision and
  Pattern Recognition}, pages 2347--2356, 2019.

\end{thebibliography}
}

\clearpage
\appendix
\section{Appendix}
%
%
This appendix complements the main manuscript and provides more qualitative results as well as questions from the user study. 
%

\subsection{More Qualitative Results}
\begin{figure*}[t]
    \includegraphics[width=0.95\linewidth]{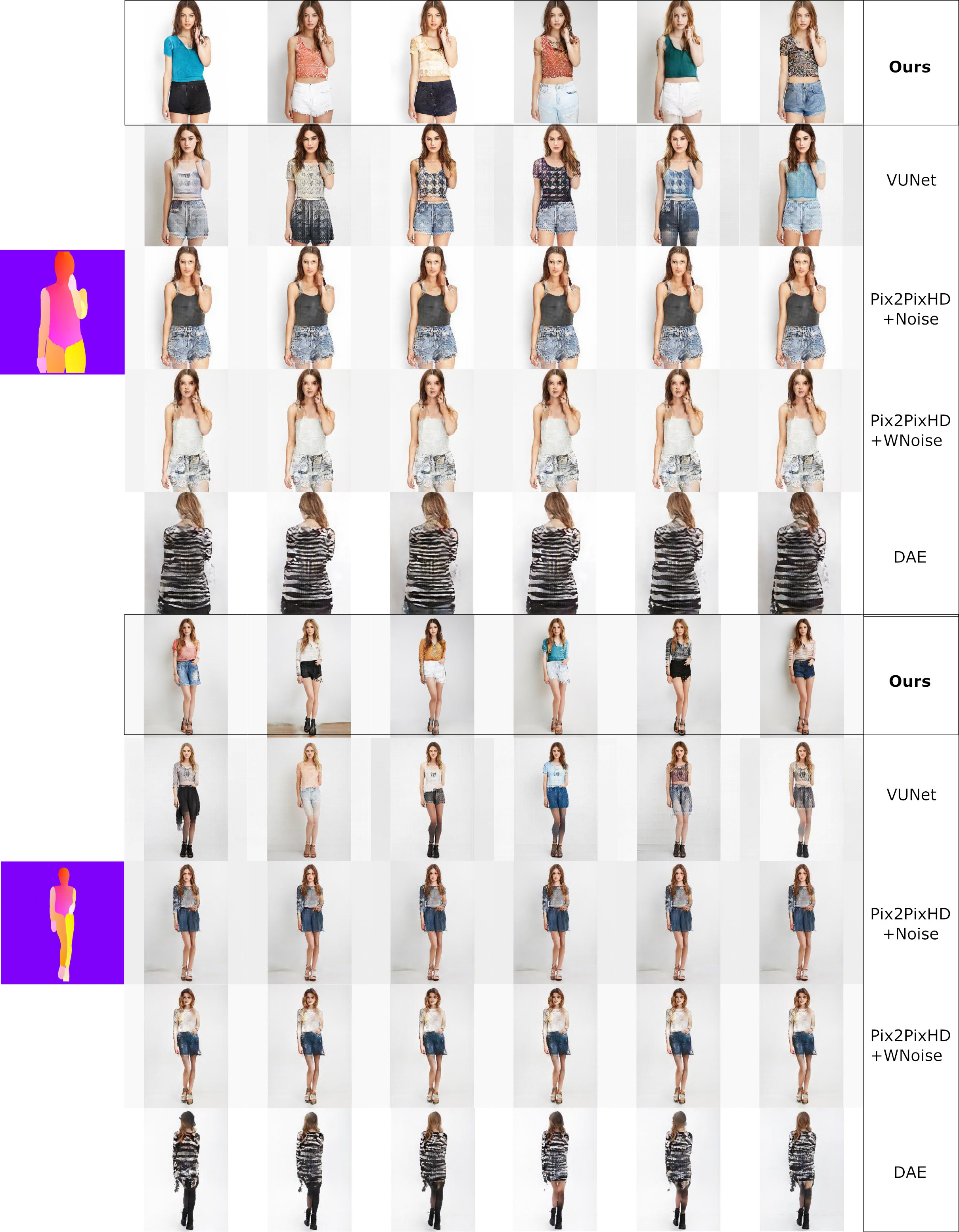}
    \caption{Results of our method for pose-guided image generation and its comparison with VUNet \cite{esser2018variational} and other baselines. The conditioning pose in the form of DensePose is shown in the left column.} 
    \label{fig:app_supp1} 
\end{figure*}

\begin{figure*}[t]
    \includegraphics[width=0.95\linewidth]{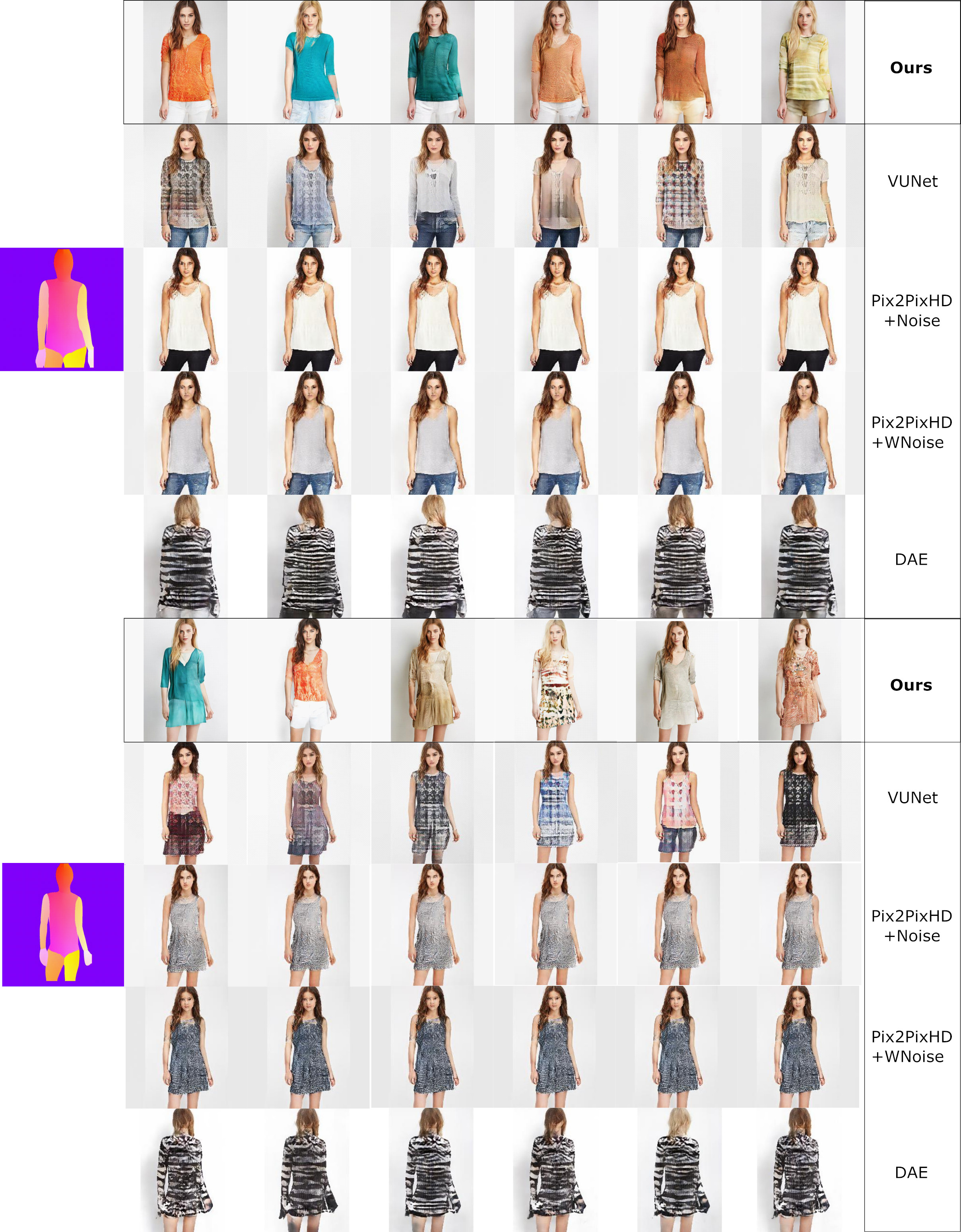}
    \caption{Results of our method for pose-guided image generation and its comparison with VUNet \cite{esser2018variational} and other baselines. The conditioning pose in the form of DensePose is shown in the left column.} 
    \label{fig:app_supp2} 
\end{figure*}

\begin{figure*}[t]
    \includegraphics[width=0.95\linewidth]{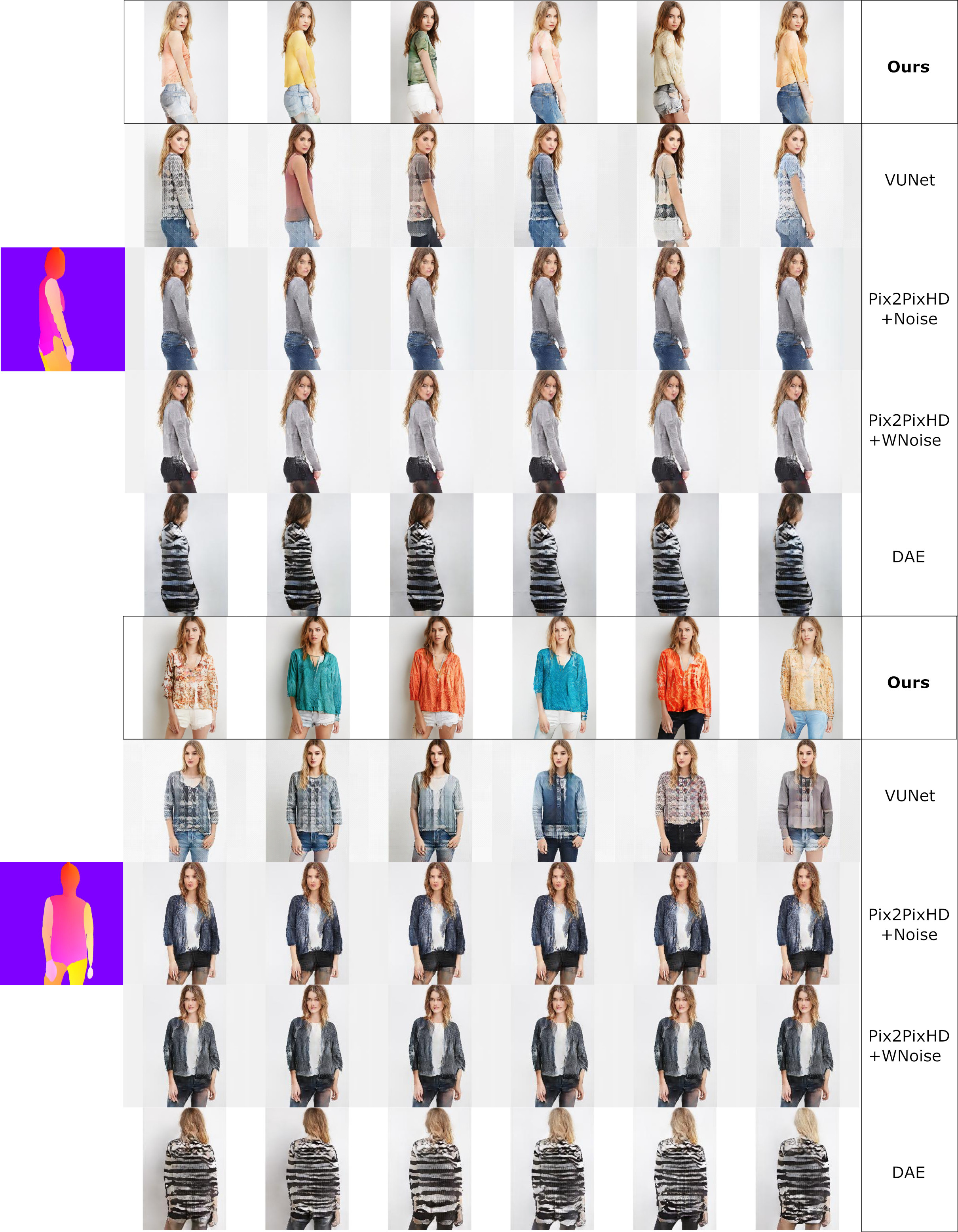}
    \caption{Results of our method for pose-guided image generation and its comparison with VUNet \cite{esser2018variational} and other baselines. The conditioning pose in the form of DensePose is shown in the left column.} 
    \label{fig:app_supp3} 
\end{figure*}

\paragraph{Appearance Sampling.}
Figs.~\ref{fig:app_supp1}, \ref{fig:app_supp2}, and \ref{fig:app_supp3} show the qualitative results of our method for pose-guided image generation and its comparison with the baselines \textit{VUNet} \cite{esser2018variational}, \textit{Pix2PixHD+Noise}, \textit{Pix2PixHD+WNoise} and \textit{DAE} (Deterministic Auto Encoder). We observe that purely noise-based baselines such as \textit{Pix2PixHD+Noise} not only failed to generate diverse output but also lacked realism. VUNet produces a diverse set of outputs but often shows spurious patterns. It also lacks realism because of its GAN-free  architecture and less powerful generator. Our approach, in contrast, produces samples of high quality than the baseline methods. 

\begin{figure*}[t]
    \includegraphics[width=\linewidth]{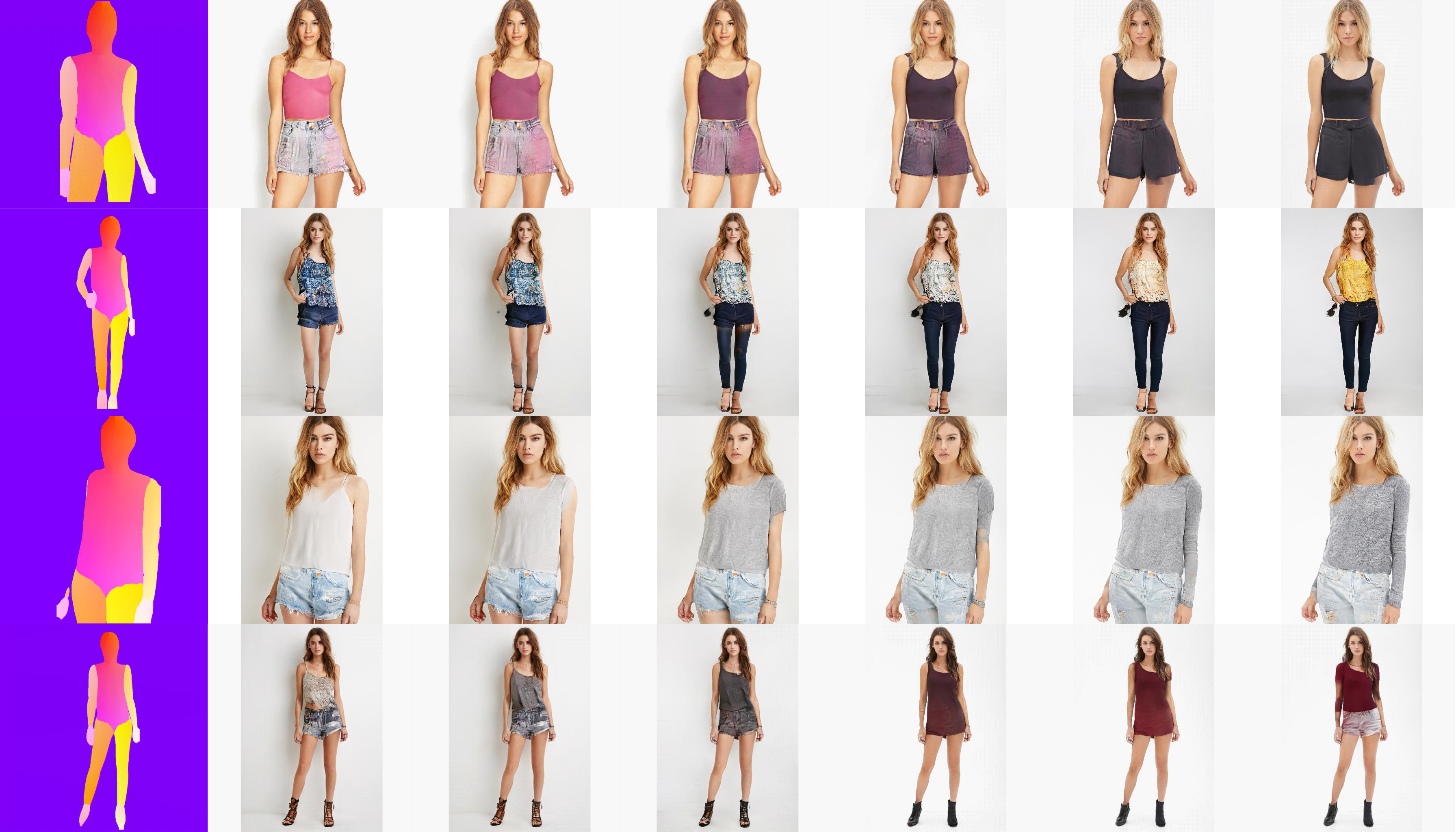}
    \caption{Generated images with interpolated appearance encodings. The conditioning pose is shown on the left.} 

    \label{fig:inter_supp3} 
\end{figure*}

\paragraph{Image Interpolation.} 
Fig.~\ref{fig:inter_supp3} shows the resulting images generated by interpolating the appearance vector between two encodings: given two encodings $\bm z_1$ and $\bm z_2$ of two different human images, we generate a human image with the interpolated encoding $\bm z$ as 
\begin{equation} 
z = z_1 t + z_2(1-t),\;t \in [0, 1]. 
\end{equation} 
We find that intermediate images show coherently dressed humans that share the properties of both input images. 

\begin{figure*}[t]
    \includegraphics[width=\linewidth]{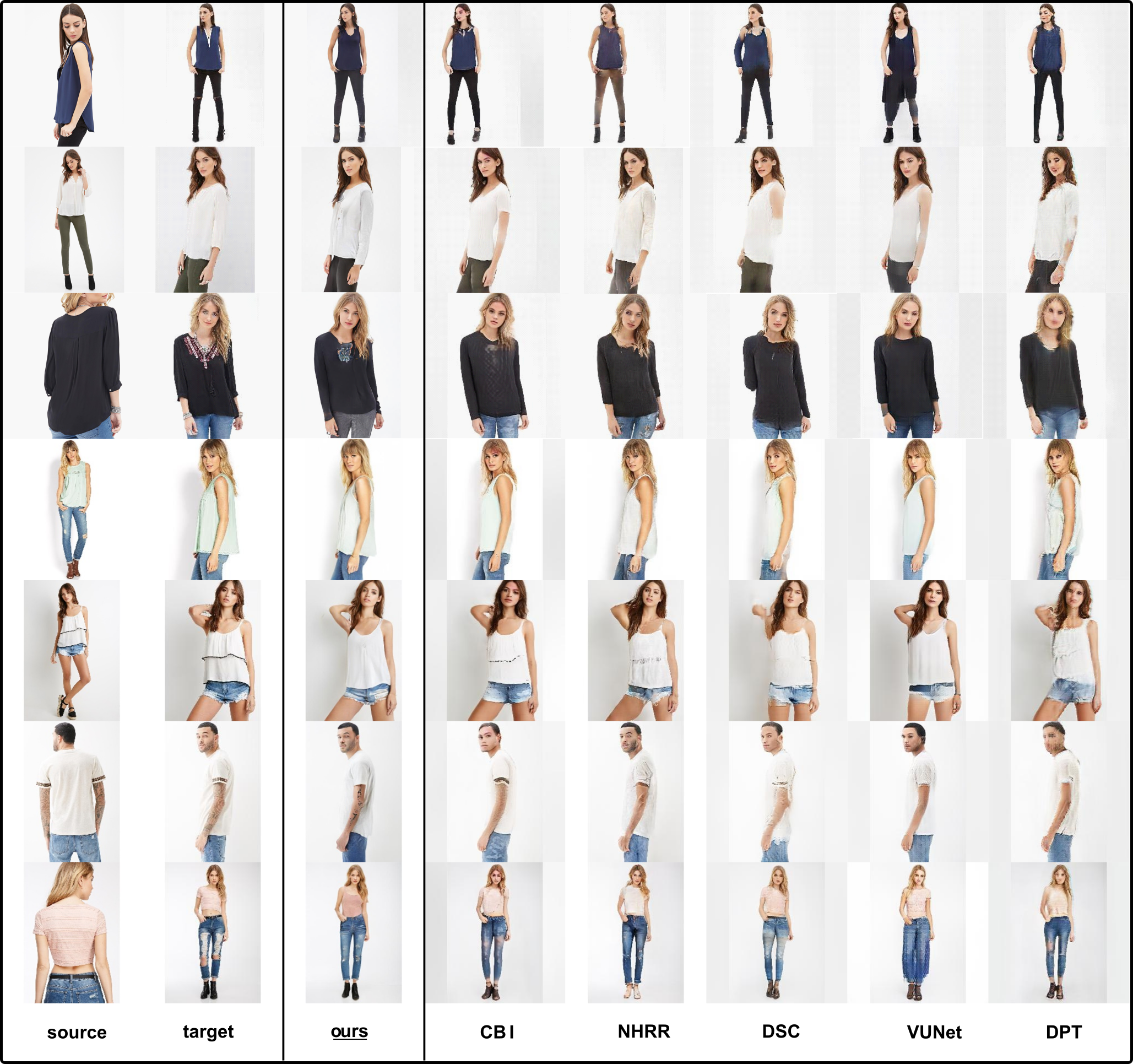}
    \caption{\textbf{Pose Transfer.} Comparison of our reconstruction+transfer results with the state-of-the-art pose transfer methods CBI \cite{Grigorev2019CoordinateBasedTI},  NHRR \cite{Sarkar2020},  DSC \cite{Siarohin2019AppearanceAP}, VUNet \cite{esser2018variational} and DPT \cite{Neverova2018}. Our HumanGAN produces more realistic renderings than the competing methods.} 
    \label{fig:transfer_supp} 
\end{figure*}

\paragraph{Pose Transfer.}
Fig.~\ref{fig:transfer_supp} shows our results for pose transfer and its comparison with the state-of-the-art methods. Our results show higher realism and are more visually pleasing compared to the other baselines. 
However, it misses some fine-scaled details in a few cases. See  Figs.~\ref{fig:user_study1} and \ref{fig:user_study2} for more samples.

\begin{figure*}[t]
\centering
    \includegraphics[width=0.8\linewidth]{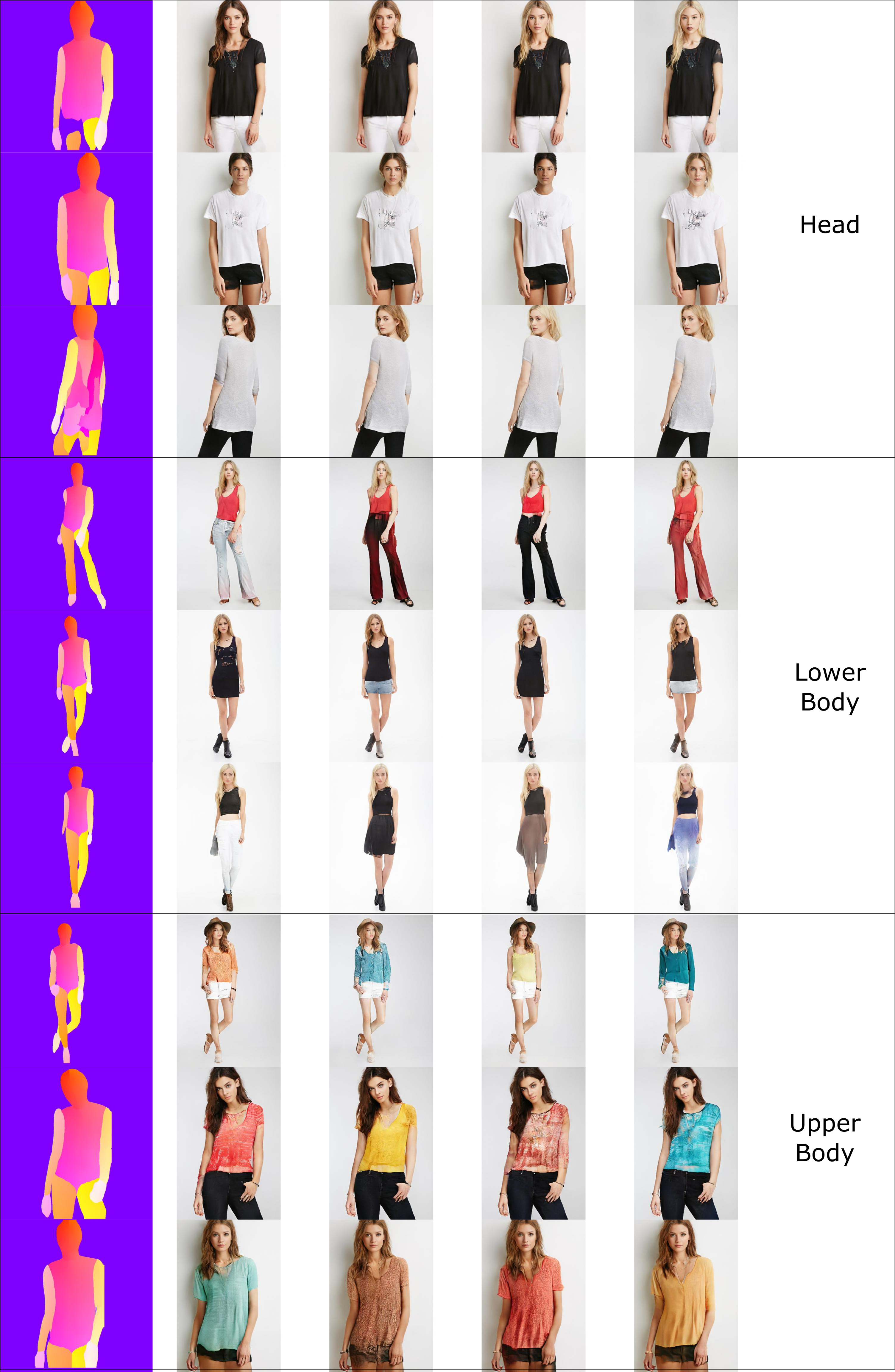}
    \caption{Our results for part sampling (head, lower body and upper body). Conditioning pose is shown in the left column.} 
    \label{fig:parts_supp} 
\end{figure*}

\begin{figure*}[t]
\centering
    \includegraphics[width=0.6\linewidth]{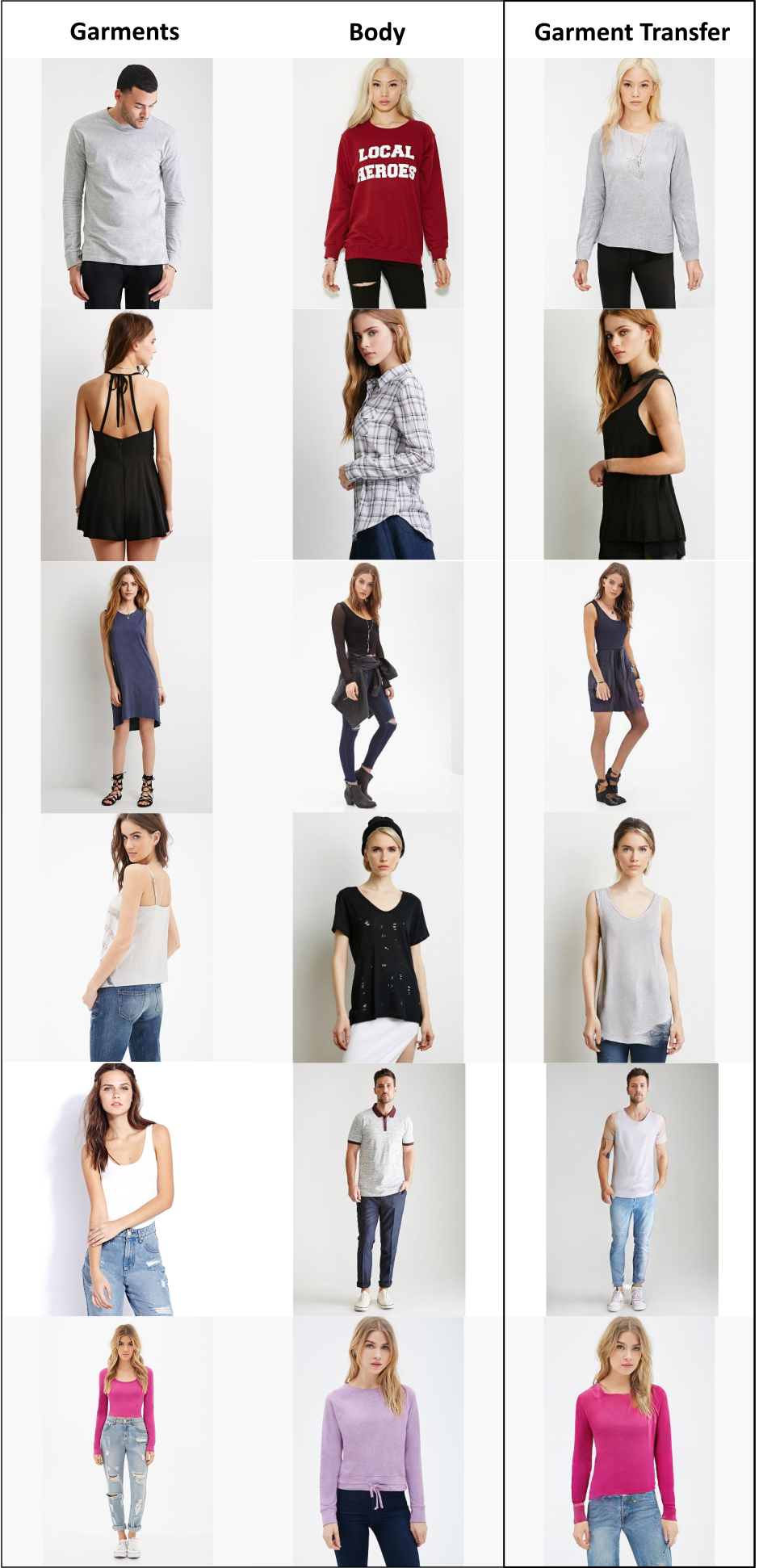}
    \caption{Our results for garment transfer.} 
    \label{fig:garments_supp} 
\end{figure*}

\paragraph{Part Sampling and Garment Transfer.}
Fig.~\ref{fig:parts_supp} shows our results for part sampling. Here, we only change the latent vectors representing a specific part (\textit{e.g.,} head). We observe that the rest of the body does not change considerably with the change in the generated image parts. However, we also observe that our method is biased towards generating realistic outputs over generating images that are highly different in the part regions but not coherent as a whole (\textit{e.g.,} sampling the head and  garments of a female will result in images with female heads). 
Fig.~\ref{fig:garments_supp} shows our garment transfer results, where we use the appearance encodings of two different images corresponding to the body and garment parts. 

\vspace{8pt}
\subsection{Comparisons to Additional Baselines} 
\label{sec:additional_baselines}
Variational autoencoders are notoriously difficult to train, and we have made several observations while developing and training our model. 
Appending noise to the conditioning DensePose image (baseline \textit{Pix2PixHD+Noise}) not only failed to generate diverse output but also lacked realism. 
Global latent vectors (instead of the part-specific) for appearance (baseline \textit{NoParts}) lacked realism as well. This section shows the results of two additional baselines (also introduced in the main manuscript):

a) \textbf{+DPCond.} This baseline conditions the DensePose image $\bm P_t$ in addition to the noise image $\bm Z_t$ to the generator. \textit{i.e.} we concatenate  $\bm P_t$ and $\bm Z_t$ channel-wise and input the resulting tensor to the generator. 

b) \textbf{+NoisePrior.} This baseline use samples from the prior $\mathcal{N}(0,1)$  (along with the encoded distribution) during the training, as recommended by Larsen \textit{et al.}~\cite{larsen2016autoencoding}. 

\begin{figure*}[t]
\centering
    \includegraphics[width=1.0\linewidth]{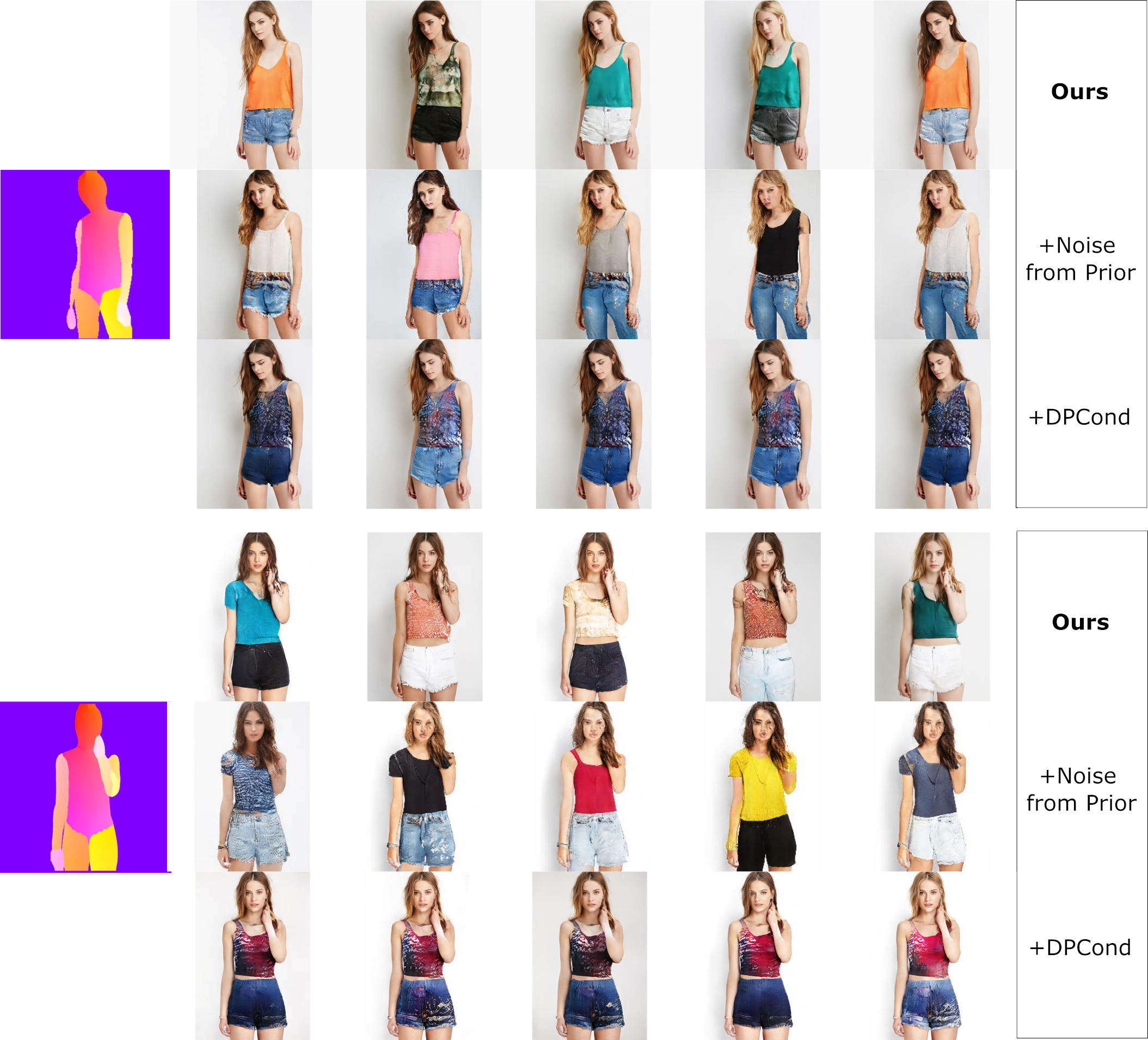}
    \caption{Comparison of our method with additional baselines. We observe that \textit{+Noise from Prior} leads to less realistic images, while \textit{+DPCond} leads to less variation during sampling. See Sec. \ref{sec:additional_baselines} for more details.} 
    \label{fig:app_discuss_supp} 
\end{figure*}

 The qualitative results are shown in Fig.~\ref{fig:app_discuss_supp}. We observe that conditioning $\bm P_t$ along with the noise image $\bm Z_t$ to the generator, resulted in fewer variations during sampling. We assume the reason to be the overpowering of the conditioning variable --- it does not let the latent vectors learn semantics. 
In our HumanGAN, we force the generator to produce output only from the warped  noise vector, thereby enforcing semantics for sampling. 
We also observe that using samples from the prior $\mathcal{N}(0,1)$ during the training leads to high variation during sampling. However, it creates highly distorted faces and other body parts.

\subsection{User Studies}
\begin{figure*}[t]
\centering
    \includegraphics[width=0.95\linewidth]{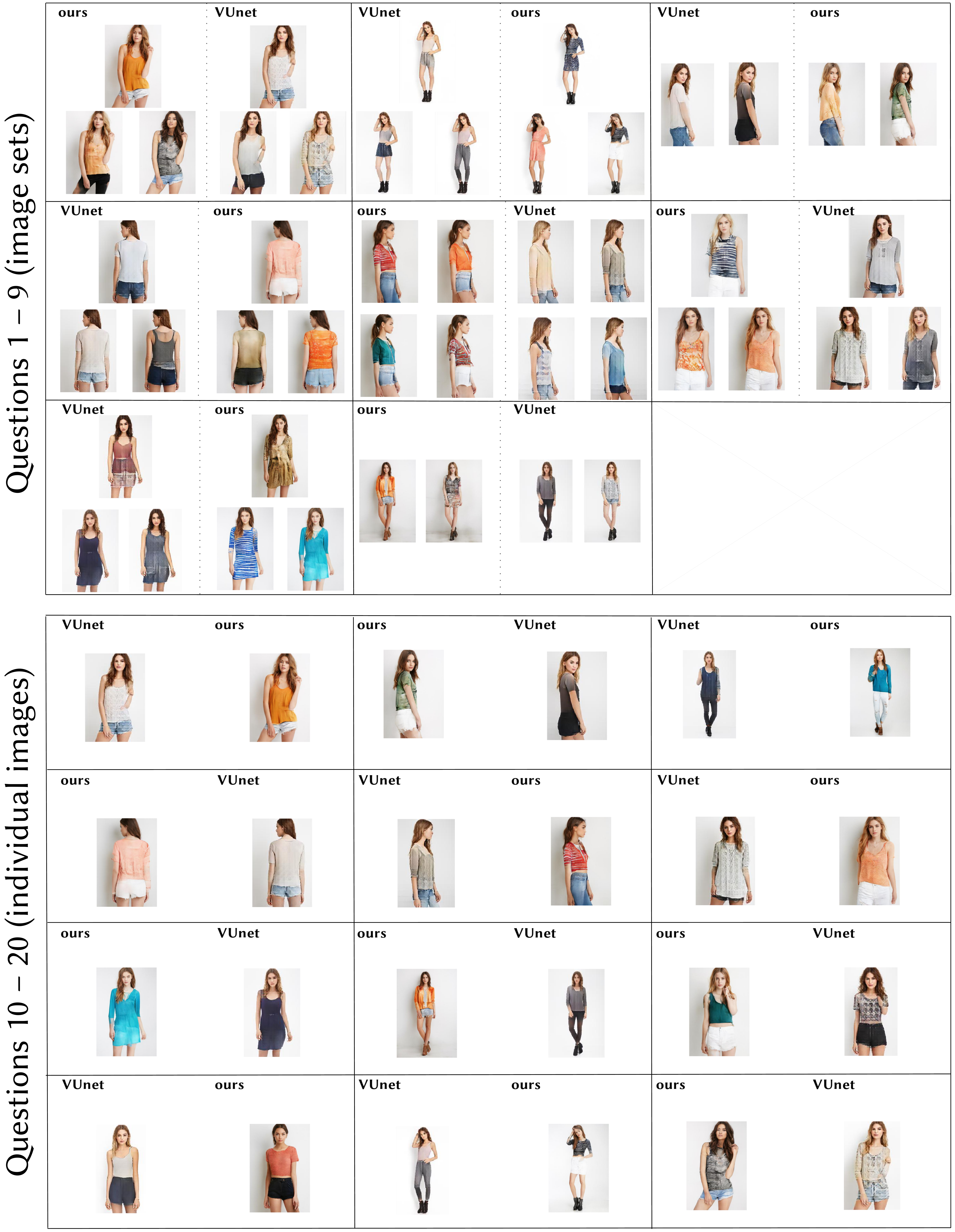}
    \caption{The samples and the sets used in the \textit{first user study} where we compare our results with VUNet \cite{esser2018variational} for  appearance sampling. The keys on the top left were not shown during the user study (they are replaced with \textit{A} and \textit{B} variants).}
    \label{fig:user_study1} 
\end{figure*}

\begin{figure*}[t]
\centering
    \includegraphics[width=\linewidth]{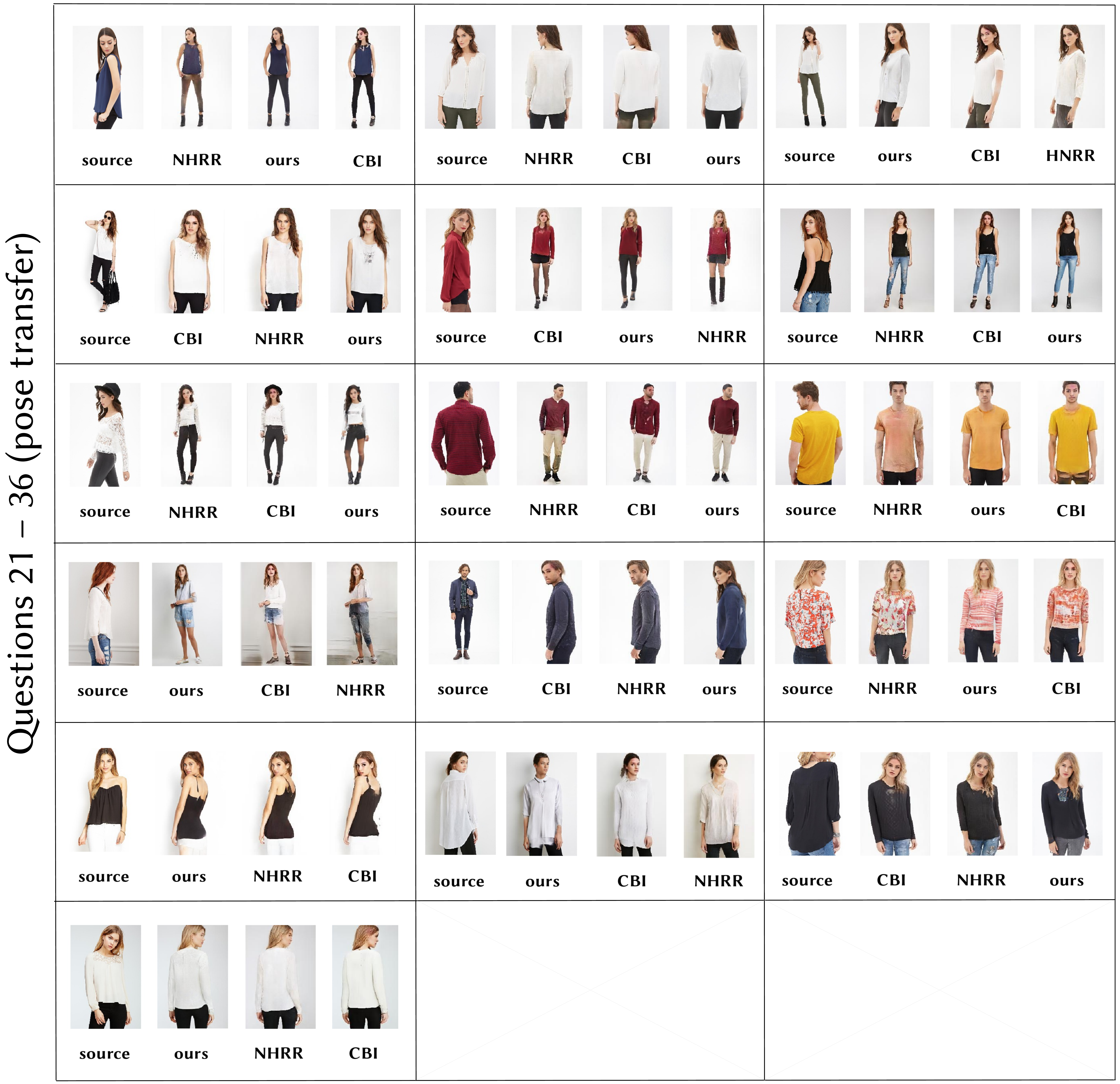}
    \caption{The samples and the sets used in the \textit{second user study} where we compare our results with CBI \cite{Grigorev2019CoordinateBasedTI} and NHRR \cite{Sarkar2020} for pose transfer. The keys on the bottom were not shown during the user study (they are replaced with \textit{A}, \textit{B} and \textit{C} variants).} 
    \label{fig:user_study2} 
\end{figure*}

In Figs.~\ref{fig:user_study1} and \ref{fig:user_study2}, we  show the list of questions used in the two user studies. 

\end{document}